\documentclass[letterpaper]{article} 
\usepackage[]{aaai2026}  
\usepackage{times}  
\usepackage{helvet}  
\usepackage{courier}  
\usepackage[hyphens]{url}  
\usepackage{graphicx} 
\usepackage{natbib}  
\usepackage{caption} 
\usepackage{algorithm}
\usepackage{algpseudocode}

\usepackage{booktabs}
\usepackage{amsmath}
\usepackage{amsfonts,amssymb}
\usepackage{tabularx}
\usepackage[table]{xcolor}

\usepackage{newfloat}
\usepackage{listings}
\DeclareCaptionStyle{ruled}{labelfont=normalfont,labelsep=colon,strut=off} 
\floatstyle{ruled}
\newfloat{listing}{tb}{lst}{}
\floatname{listing}{Listing}
\title{Policy Search, Retrieval, and Composition via Task Similarity in Collaborative Agentic Systems}
\author{
    Saptarshi Nath\textsuperscript{\rm 1},
    Christos Peridis\textsuperscript{\rm 1},
    Eseoghene Benjamin\textsuperscript{\rm 4},
    Xinran Liu\textsuperscript{\rm 2},
    Soheil Kolouri\textsuperscript{\rm 2},
    Peter Kinnell\textsuperscript{\rm 1},
    Zexin Li\textsuperscript{\rm 3},
    Cong Liu\textsuperscript{\rm 3},
    Shirin Dora\textsuperscript{\rm 1},
    Andrea Soltoggio\textsuperscript{\rm 1}
}
\affiliations{
    \textsuperscript{\rm 1}Department of Computer Science, Loughborough University, Loughborough, UK\\
    \textsuperscript{\rm 2}Department of Electrical Engineering and Computer Science, Vanderbilt University, Nashville, USA\\
    \textsuperscript{\rm 3}Department of Electrical Engineering and Computer Engineering, University of California, Riverside, USA\\
    \textsuperscript{\rm 4}Alan Turing Institute, London, UK\\

    \{s.nath, c.peridis, p.kinnell, s.dora, a.soltoggio\}@lboro.ac.uk, \{xinran.liu, soheil.kolouri\}@vanderbilt.edu, \{zli536, congl\}@ucr.edu, ebenjamin@turing.ac.uk
}

\begin{document}

\maketitle

\begin{abstract}
    Agentic AI aims to create systems that set their own goals, adapt proactively to change, and refine behavior through continuous experience. Recent advances suggest that, when facing multiple and unforeseen tasks, agents could benefit from sharing machine-learned knowledge and reusing policies that have already been fully or partially learned by other agents. However, how to query, select, and retrieve policies from a pool of agents, and how to integrate such policies remains a largely unexplored area.  This study explores how an agent decides what knowledge to select, from whom, and when and how to integrate it in its own policy in order to accelerate its own learning. The proposed algorithm, \emph{Modular Sharing and Composition in Collective Learning} (MOSAIC), improves learning in agentic collectives by combining (1) knowledge selection using performance signals and cosine similarity on Wasserstein task embeddings, (2) modular and transferable neural representations via masks, and (3)  policy integration, composition and fine-tuning. MOSAIC outperforms isolated learners and global sharing approaches in both learning speed and overall performance, and in some cases solves tasks that isolated agents cannot. The results also demonstrate that selective, goal-driven reuse leads to less susceptibility to task interference. We also observe the emergence of self-organization, where agents solving simpler tasks accelerate the learning of harder ones through shared knowledge.
\end{abstract}

\begin{links}
    \link{Code}{https://github.com/DMIU-ShELL/MOSAIC}
\end{links}

\section{Introduction}
The ability to solve complex, open-ended problems is an increasingly desirable objective as agentic AI principles become central in AI architectures. In open-ended real-world settings, these systems face challenges in generalizing and adapting to an ever-evolving stream of data and tasks  \cite{delangecl2022survey, parisi2019continual}. Although new lifelong learning algorithms \cite{kudithipudi2022biological} are being introduced to provide continual evolution and adaptation, when agents learn in isolation, they can only benefit from their limited experiences, unlike humans who benefit from collaboration and sharing of experiences and skills. Recent studies suggest that a shift towards more collaborative types of learning, where agents selectively share and reuse modular knowledge from other agents, is key to reducing redundant learning, accelerating adaptation, and improving robustness \cite{TaraleDistributedLearning, soltoggio2024nature, DARPA:ShELL}.

In the context of self-adaptation, reinforcement signals provide an effective tool to achieve continuous improvement in a wide range of domains \cite{sutton2018reinforcement, Jaderberg2016ReinforcementTasks}. In addition, task similarities have been exploited to benefit learning in areas such as transfer learning and multi-task RL \cite{Hendawy2023Multi-TaskExperts, sun2022paco, D'Eramo2020Sharing}. Lifelong learning approaches also leverage task similarity when previously learned tasks benefit the learning of subsequent tasks, with an advantage often named \emph{forward transfer}  \cite{ben2022lifelong, kirkpatrick2017ewc}. However, many approaches to sharing knowledge, e.g., distributed RL \cite{sartoretti2019distributed} or Federated Learning \cite{yoon2021fedweit}, often assume some form of centralization and uniformity of tasks, which is ill-suited to the requirements of agentic AI where each agent acts independently, asynchronously, and likely on a large variety of different tasks. Recent studies have begun to investigate the sharing and reuse of policies in evolving multi-agent contexts \cite{nath2023L2D2-C, TaraleDistributedLearning, Gerstgrasser2023SelectivelyLearning}. These developments reflect a growing recognition that scalable AI systems increasingly depend on self-directed selective collaboration among autonomous learners.

Knowledge and skill sharing among a set of independent and autonomous learners, while conceptually appealing, presents significant challenges. In particular, open research questions reflect the problems of how to identify what knowledge to acquire, from whom to acquire it, how such knowledge is best represented, and how to integrate it to manifest robust lifelong learning properties \cite{DARPA:ShELL, soltoggio2024nature}. Furthermore, it is still unclear to what extent collaborative learning offers an advantage over isolated or centralized learning.

To address these questions, this study introduces \emph{Modular Sharing and Composition in Collective Learning} (MOSAIC), which describes how agents can select appropriate knowledge from peers with the aim of implementing policy composition and reuse. Problem analysis and the following ablation studies indicate that essential components of such a system are (1) knowledge selection through task similarity, (2) modular and transferable neural representations, and (3) policy integration, composition, and fine-tuning.

Specific algorithmic choices in MOSAIC to implement such components are Wasserstein embeddings for task similarity, transferable binary network masks for sharing, and learnable linear combinations of those to achieve integration and fine tuning. In principle, analogous methods could be used to adapt MOSAIC to a variety of domains, e.g., LoRA-based modular composition for foundation models and LLMs \cite{hu2022lora, He2021TowardsLearning} and embedding methods or descriptors to express task alignment \cite{Achille_2019_ICCV, pmlr-v80-grover18a, pmlr-v97-rakelly19a}.

Simulations indicate that selectively acquiring relevant knowledge is essential to effectively reuse policies. The composition of policies is best achieved with a similarity-driven and reward-aware weighting. The results show how communicating MOSAIC agents learn faster than isolated learners by reusing knowledge from peers on similar tasks. In some cases, communicating agents can solve tasks that they could not learn alone. In addition, the analysis reveals an implicit self-organization, where agents discover and exploit curriculum structures in the available knowledge, resulting in agents building on skills hierarchically, from simpler tasks to learn more complex ones.

In summary, MOSAIC supports the hypothesis that task-relevant policy transfer and reuse, guided by agent-centered optimization objectives, is feasible and  effective with advantages over learning in isolation or sharing global parameters, which could lead to task interference. To the best of our knowledge, this is the first study that combines modular and transferable task-specific knowledge with similarity-based selection and integration. Such a combination of components is sufficient to significantly improve sample efficiency and learning through collective knowledge sharing.

\section{Related Work} \label{sec:related_work}
Recent work in modular knowledge reuse for multi-agent RL explores various approaches to enable efficient knowledge transfer. Federated learning methods  \cite{yoon2021fedweit} communicate task-specific masks or binary representations. These approaches lack principled task similarity metrics for selective reuse. MOSAIC addresses this gap through Wasserstein-based similarity measures that guide knowledge selection based on performance signals.

Several frameworks allow parameter-efficient knowledge sharing through masking \cite{ge2023lightweight, nath2023L2D2-C, Gerstgrasser2023SelectivelyLearning}, prototype aggregation \cite{Tan2021FedProto:Clients}, or decentralized training \cite{Douillard2024DiPaCo:Composition, JaghouarOpenDiLoCo:Training}. Communication-efficient methods reduce information exchange through knowledge preservation and selective sharing \cite{mcmahan2017communication}. These approaches rely on predefined sharing protocols or require synchronous updates. MOSAIC's binary mask representation enables asynchronous, selective reuse and maintains modularity while minimizing communication overhead.

Policy composition methods like PaCo \cite{sun2022paco} interpolate within shared subspaces. These methods lack mechanisms for similarity-driven integration. Peer-to-peer frameworks \cite{TaraleDistributedLearning, javier2022collaborative} and model-based approaches \cite{Jiang2021ModelEnvironment} enable decentralized coordination. These frameworks do not combine selection with modular representations. MOSAIC combines Wasserstein-based task similarity with binary mask modularity and similarity-driven linear policy combinations. Further discussion of related works is provided in Appendix~\ref{sec:extended_related_work}.

\section{Methodology}
The following section describes the components of the MOSAIC algorithm. A high level illustration is provided in Figure \ref{fig:approach}. Pseudocode can be found in the Appendix~\ref{sec:implementation_details}. 

\subsection{Policy Representations}
\label{sec:modular}
Isolating task-specific knowledge in compact representations \cite{pmlr-v87-alet18a} is a key to making policy transfer easier and more lightweight across agents. Lifelong learning parameter isolation approaches can achieve this goal \cite{rusu2016progressive, mallya2018packnet, wortsman2020supermasks, ben2022lifelong}.

This study adopts neural network masks \cite{ben2022lifelong, wortsman2020supermasks} that have shown strong results in lifelong reinforcement learning and supervised learning settings, and supports composition through linear combinations of mask modules. PPO is used for the experiments in this paper; however, the proposed modular composition mechanism is, in principle, agnostic to the choice of RL algorithm.
\begin{figure*}
    \centering\includegraphics[width=1.0\textwidth]{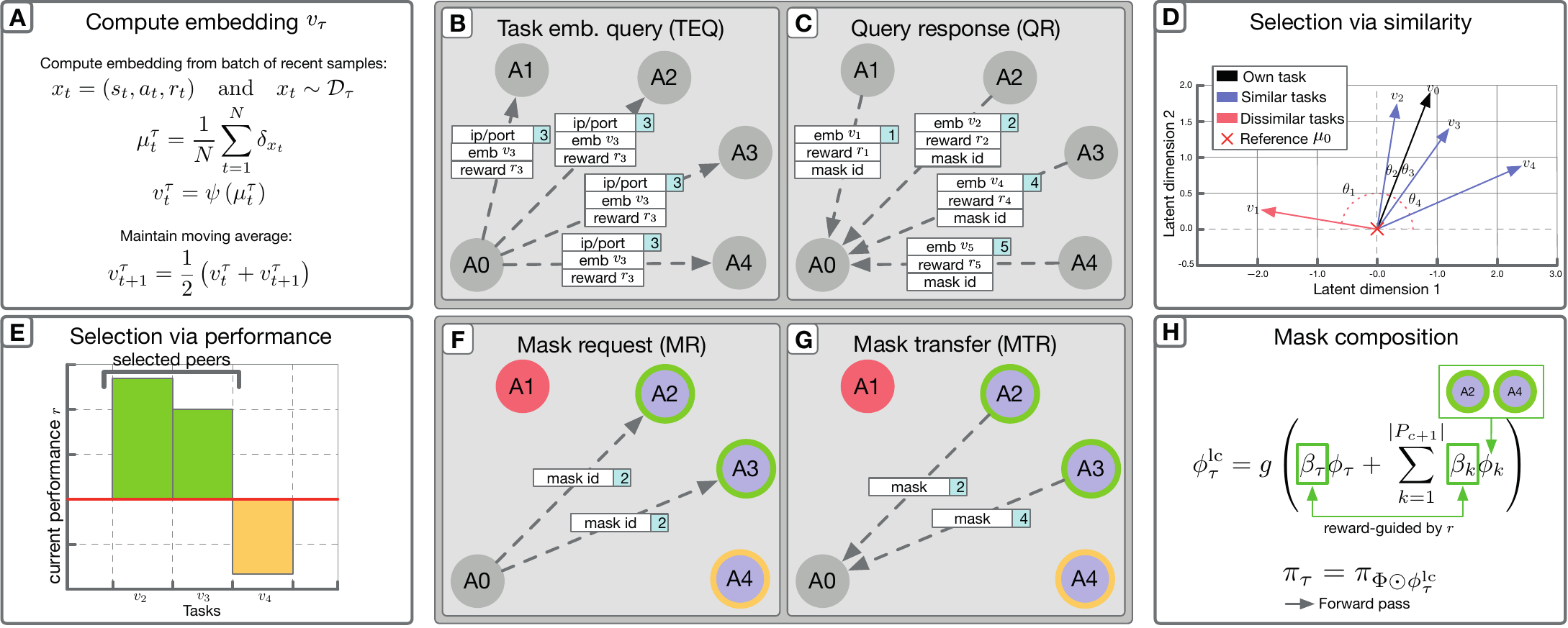}
     \caption{High-level illustration of the main MOSAIC  algorithmic steps. (A) A Wasserstein task embedding \(v_\tau\) is computed from a SAR batch by Agent A0 (representing any agent in the collective). (B) Periodically, the agent A0 broadcasts a task embedding query (TEQ) to all known peers. (C) Peers send back  a query response (QR) that contains their task's \(v_\tau, r\), and the corresponding mask ID; (D) Agent A0 selects relevant embeddings using cosine similarity on the Wasserstein embeddings. (E) A0 further selects relevant embeddings using \ref{eq:criterion_align} and \ref{eq:criterion_perf}. (F) A0 sends mask requests (MR). (G) The contacted agents respond by sending the requested mask through a mask transfer (MTR). (H) Incoming masks from A2 and A3 are incorporated into A0. The training of the agent's policy occurs in parallel (not represented in the figure).}
    \label{fig:approach}
\end{figure*}
Mask representations rely on a frozen backbone network \(\Phi\) that is homogeneous across all agents. Each policy \(\pi_\tau\) is implicitly parameterized through a sparse mask \(\phi_\tau\), such that \(\pi_\tau=\pi_{\Phi\odot g(\phi_\tau)}\), where \(\odot\) denotes the Hadamard product, and \(g(\cdot)\) is a binarization function applied during the forward pass. The mask \(\phi_\tau\) consists of a set of real-valued score vectors, one per layer of the shared backbone network \(\Phi\), which select a sparse task-specific subnetwork from \(\Phi\) without modifying its parameters. Scores are binarized during the forward pass to select a discrete subnetwork from the backbone, given by the thresholding function,
\begin{equation} \label{eq:binarization}
    g(\phi_\tau)_\ell =
    \begin{cases}
        1 & \text{if } (\phi_\tau)_\ell > \epsilon \\
        0 & \text{otherwise}
    \end{cases}, \quad \text{with } \epsilon = 0 \quad,
\end{equation}
where \(g(\phi_\tau)\) denotes the complete binary mask and \(\ell\) indexes the layers of the backbone network \(\Phi\). During the backward pass, gradients are computed with respect to the real-valued mask parameters \(\phi_\tau\), and updates are performed using the straight-through estimator (STE) \cite{wortsman2020supermasks} to enable learning through the non-differentiable binarization step.

\subsection{Wasserstein Task Embeddings for Online Reinforcement Learning}
\label{sec:wasserstein_sim}
To identify suitable policies for a given task, MOSAIC
extends \citet{liu2022wasserstein} to RL by computing embeddings over batches of state-action-reward (SAR) data collected online (see Figure~\ref{fig:approach}(A)). For task \(\tau\), the agent maintains an empirical task distribution
\begin{equation}\label{eq:embedding_distribution}
    \mu_\tau = \frac{1}{N} \sum_{t=1}^{N} \delta_{x_t}\in \mathcal{P}(\mathbb{R}^d),
\end{equation}
where \(x_t = (s_t, a_t, r_t)\) and \(x_t \sim \mathcal{D}_\tau\). 
Here, \(\mathcal{D}_\tau\) denotes the replay buffer for \(\tau\), and each 
\(x_t \in \mathbb{R}^d\) is a state-action-reward (SAR) tuple. 
The tuple dimensionality is \(d = d_s + d_a + d_r\), 
where \(d_s\), \(d_a\), and \(d_r\) are the dimensions of the state, action, and scalar reward, respectively. We use \(N=128\) samples in our experiments. \(\delta_{x_t}\) denotes the Dirac measure centered on the sample \(x_t\). A fixed synthetic reference distribution \(\mu_0\) is defined as
\begin{equation}\label{eq:reference_distribution}
    \mu_0 = \frac{1}{M}\sum_{m=1}^M \delta_{x_m^0},
\end{equation}
where \(x_m^0 \sim \text{Uniform}(-1,1)^d\) is sampled once during initialization and fixed thereafter. We fix \(M=50\) as the number of reference points in the synthetic distribution. This shared reference is used across all agents and enables them to align their embeddings in a shared latent space without centralized training or supervision. The 2-Wasserstein distance between \(\mu_\tau\) and \(\mu_0\) is computed by solving the optimal transport problem,
\begin{equation}
\begin{aligned}
\gamma^* &= \arg\min_{\gamma \in \mathbb{R}^{N\times M}}
\sum_{t=1}^N \sum_{m=1}^M \gamma_{tm} \|x_t - x^0_m\|^2 \\
&\quad \text{s.t. }
\gamma_{tm} \ge 0,\;
\gamma \mathbf{1}_M = \tfrac{1}{N} \mathbf{1}_N,\;
\gamma^\top \mathbf{1}_N = \tfrac{1}{M} \mathbf{1}_M,
\end{aligned}
\end{equation}
where \(\mathbf{1}_N\) and \(\mathbf{1}_M\) are vectors of ones in \(\mathbb{R}^N\) and \(\mathbb{R}^M\), respectively. The Wasserstein task embedding operator \(\psi\) maps the task distribution \(\mu_\tau\) to a vector \(v_\tau \in \mathbb{R}^{M\times d}\) via the barycenter projection of \(\gamma^*\) on \(\{x_t\}_{t=1}^N\),
\begin{equation} \label{eq:embedding_mapping}
    v_\tau = \psi(\mu_\tau) = \left[ \sum_{t=1}^N \gamma_{tm}^* x_t \right]_{m=1}^M.
\end{equation}
This mapping places each task \(\tau\) in a shared latent space, where metrics can be used to quantify task relationships. A moving average is maintained to smooth fluctuations as the agent's policy evolves, \(v_{t+1}^\tau=\frac{1}{2}(v_t^\tau+v_{t+1}^\tau)\).

\subsection{Knowledge Selection and Sharing}
\label{sec:selection}
Agents communicate by maintaining a list of IP addresses and ports for other agents, which allows them to be located anywhere on the Internet. Each agent \(i \in \mathcal{I}\) can communicate directly with all other agents \(j \in \mathcal{I},\ j \ne i\) at any time.
\paragraph{Phase 1: Embedding queries}
At any time, an agent \(i\) can initiate a task embedding query (TEQ) by broadcasting its current task embedding \(v_i\), iteration performance \(\overline{r}_i\), and its IP/port (see Figure \ref{fig:approach}(B)(C)). Upon receiving the TEQ, each peer agent \(j\ne i\) responds with its most recent task embedding \(v_j\) and associated performance \(\overline{r}_j\).

\paragraph{Phase 2: Policy selection.}
Once the agent has received query responses (QR) (see Figure~\ref{fig:approach}(F)(G)) with relevant embeddings from other agents, it computes the cosine similarity between its own embedding and those of each peer,
\begin{equation}
    \text{cos}(v_i, v_j) = \frac{ \langle \text{vec}(v_i), \text{vec}(v_j) \rangle }{ | \text{vec}(v_i) | \cdot | \text{vec}(v_j) | }, \quad \text{cos} \in [-1, 1].
\end{equation}
To identify useful policies, agent \(i\) applies two heuristic filters based on similarity and performance:
\begin{align}
    \mathbb{I}_{\text{align}}(i, j) &= 
    \begin{cases}
        1 & \text{if } \text{cos}(v_i, v_j) > \theta, \\
        0 & \text{otherwise}
    \end{cases} \tag{Criterion 1} \label{eq:criterion_align} \\
    \mathbb{I}_{\text{perf}}(i, j) &= 
    \begin{cases}
        1 & \text{if } \overline{r}_j > \overline{r}_i, \\
        0 & \text{otherwise}
    \end{cases} \tag{Criterion 2} \label{eq:criterion_perf}
\end{align}
with \(\theta=0.5\) in all experiments. The first criterion promotes semantic alignment between tasks, revealing latent similarities that can lead to effective transfer (see Section~\ref{sec:sim_transfer}). The second ensures that agents only acquire masks from peers whose performance exceeds their own, avoiding noisy or under-trained policies that can degrade performance. Criterion 2 guards against the frequent false positives seen when comparing poorly trained policies that yield deceptively similar embeddings.

The two phases and the selection criteria also result in a bandwidth-efficient approach, since policies are transferred only when considered potentially useful. The peer masks that pass both criteria are stored as the set \(P_{c+1}=\{\phi_1,...,\phi_K\}\), where each \(\phi_k\) denotes a mask received from a selected peer.

\subsection{Knowledge Composition and Fine Tuning }
\label{sec:knowledge_comp}
Masks received from other agents meeting the two selection criteria above have been pretrained on tasks that are likely related to or share similarities with the agent's current task. As previously shown in \citet{ben2022lifelong}, a linear combination of masks, weighted with trainable parameters \(\beta\), can lead to beneficial policy search in RL. MOSAIC exploits this idea by combining policies that have been trained on the local agent, plus those acquired from other agents. In this study, the \(\beta\) values are computed using softmax on a set of beta parameters, optimized in log-space, \( \tilde{\beta} \in \mathbb{R}^{1 + |P_c|} \), where \(P_c\) is the number of masks acquired during a communication event \(c\). During the forward pass, the agent constructs the linearly combined, binarized mask,
\begin{equation}\label{eq:linearcombination}
    \phi_\tau^\text{lc} = g\left(\beta_\tau \phi_\tau + \sum_{k=1}^{|P_{c+1}|}\beta_k\phi_k \right),
\end{equation}
where \( \phi_\tau \) is the agent's own task mask, and \( \phi_k \) is the \( k \)-th peer mask. The parameters \( \beta_\tau \) and \( \beta_k \) are the softmax-normalized values derived from \( \tilde{\beta} \), for \(\phi_\tau\) and \(\phi_k\), respectively. \(|P_{c+1}|\) is the number of masks acquired in the next communication event. The resulting binary mask, \( \phi_\tau^{\text{lc}} \), is then used to modulate the backbone parameters, which gives the final policy \( \pi_\tau \). As training continues on the local task, the agent updates the \(\beta\) parameters and the real value \(\phi_\tau\) via backpropagation (keeping all \(\phi_k\) masks fixed) thus fine tuning the overall resulting policy determined by \(\phi_\tau^\text{lc}\).

\paragraph{Reward-Guided Initialization (RGI).} The trainable parameters, \(\beta\), enable gradient descent to determine which policies are most useful for the current task. The initial weighting as the masks are received is given by,
\begin{equation}\label{eq:reward-guided-init}
    \beta_{\tau} = 0.5 + 0.5\overline{r}, \quad \beta_k = \frac{0.5(1 - \overline{r})}{|P_{c+1}|}\quad,
\end{equation}
where \(\overline{r} \in [0,1]\) is the agent's normalized return from the last iteration. This scheme biases low-performing agents toward external knowledge and high-performing agents toward their own policies. Ablation studies show that such  an initialization provides a strong advantage, even though \(\beta\) is later tuned by gradient descent.

Before integrating a new set of peer masks \(P_{c+1}\), the agent consolidates its current task mask \(\phi_\tau\) with the previous peer masks \(P_c\) using a weighted linear combination,
\begin{equation}
    \phi_\tau \gets \beta_\tau\phi_\tau + \sum_{k=1}^{|P_c|}\beta_k\phi_k\quad.
    \label{eq:consolidation}
\end{equation}
The consolidation collapses multiple masks into one without affecting the policy, managing memory scalability.

\section{Experiments}
\label{sec:experiments}
MOSAIC is evaluated on three sparse-reward reinforcement learning benchmarks.  The CT-graph \cite{soltoggio2019ctgraph} and MiniHack MultiRoom \cite{Samvelyan2021MiniHackResearch} are used to assess the advantage of communicating agents versus isolated agents. The MiniGrid Crossing \cite{gym_minigrid} is used to assess the performance of MOSAIC against the chosen baselines. Each benchmark includes tasks with similar, dissimilar, and interfering tasks to assess whether MOSAIC can identify and leverage the available task similarities, while avoiding interfering policies. Each agent is assigned a unique task and trained in parallel with other agents. PPO \cite{schulman2017ppo} was used in all experiments, with an FCN for MiniGrid and CT-graph, and a CNN for MiniHack. Individual results are shown in Appendix~\ref{sec:appendix_experiments}, Figures~\ref{fig:mctgraph_individual_expanded}, \ref{fig:minihack_individual}, and \ref{fig:minigrid_individual}. Details on computing infrastructure, hyperparameters, architectures, and libraries are reported in the Appendix~\ref{sec:implementation_details}, Tables~\ref{tbl:compute_resources}, \ref{tbl:all_hyperparameters}, \ref{tbl:network_architectures}, and \ref{tbl:software_environment}. Appendix~\ref{sec:significance}, Table~\ref{tbl:sig-test} reports significance testing.

\subsection{Image Sequence Learning}
\label{sec:ct-graph}
The image sequence learning (ISL) benchmark, implemented with the Configurable Tree Graph (CT-Graph) environment \cite{soltoggio2019ctgraph}, consists of procedurally generated tree navigation problems where each node is an RL state encoded as an image. We define a curriculum of 28 tasks organized into four independent image sets, each containing seven related tasks of increasing tree depth (2–8). This configuration produces four distinct and unrelated task groups with internal hierarchies of difficulty, from easy (depth 2) to very hard (depth 8). Sparse rewards and exponential branching make the benchmark challenging, with reward probabilities as low as \(\approx 7.74 \times 10^{-9}\) for depth-8 tasks \cite{soltoggio2019ctgraph}.

Results in Figure~\ref{fig:ctgraph_minihack_results}(A) show that MOSAIC achieves significantly faster learning and broader task coverage, outperforming MOSAIC-NoComm (no sharing among agents) by 2.7\(\times\), reaching a maximum total return of 26.0 across all 28 tasks. MOSAIC reaches 50\% performance (total return of 14) in 37 iterations (18,944 steps). MOSAIC-NoComm plateaus at a maximum of 9.6. On average, MOSAIC-NoComm fails on 18 tasks, whereas MOSAIC fails on only 2.

\subsection{MiniHack MultiRoom}
MiniHack is a grid-based navigation task with sparse rewards and pixel observations. Agents must traverse connected rooms to reach a final goal. We evaluate MOSAIC on 14 tasks grouped into two difficulty clusters: room sizes of \(4\times4\) and \(6\times6\). Each level adds a room connected by a closed but unlocked door. Agents receive +1 for task completion and -0.01 for collisions. Task layouts are randomized every episode to enable learning of behavioral strategies as opposed to trajectories.
\begin{figure}
    \centering
    \includegraphics[width=1.0\columnwidth]{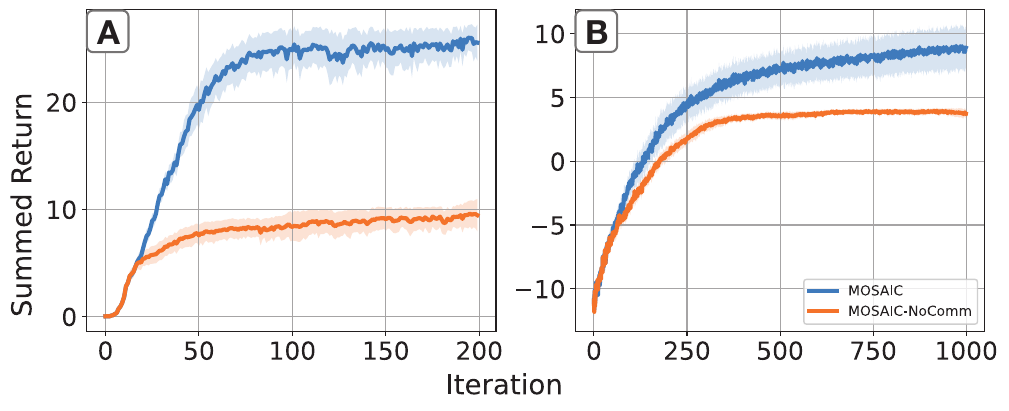}
    \caption{Performance of communicating MOSAIC agents versus isolated agents on the same tasks. (A) Image sequence learning, 28 tasks, five seeds/task: average of 140 runs with 95\% confidence intervals \cite{colas2018many}. (B) MiniHack Multiroom, 14 tasks, five seeds/task: average of 70 runs with 95\% confidence intervals. MOSAIC agents achieve relative gains of 170.8\% and 128.2\% over the isolated baseline in the image sequence and MiniHack benchmarks, respectively.}
    \label{fig:ctgraph_minihack_results}
\end{figure}
As shown in Figure~\ref{fig:ctgraph_minihack_results}(B), MOSAIC agents reach zero reward by iteration 132 (270,336 steps) versus iteration 176 (360,448 steps) for MOSAIC-NoComm, which is a 25\% reduction in the required samples. Over the full run, MOSAIC attains a maximum total return of 9.04 across all tasks, compared to 3.96 for MOSAIC-NoComm.

\subsection{MiniGrid Crossing}
\begin{figure}
    \centering
    \includegraphics[width=1.0\columnwidth]{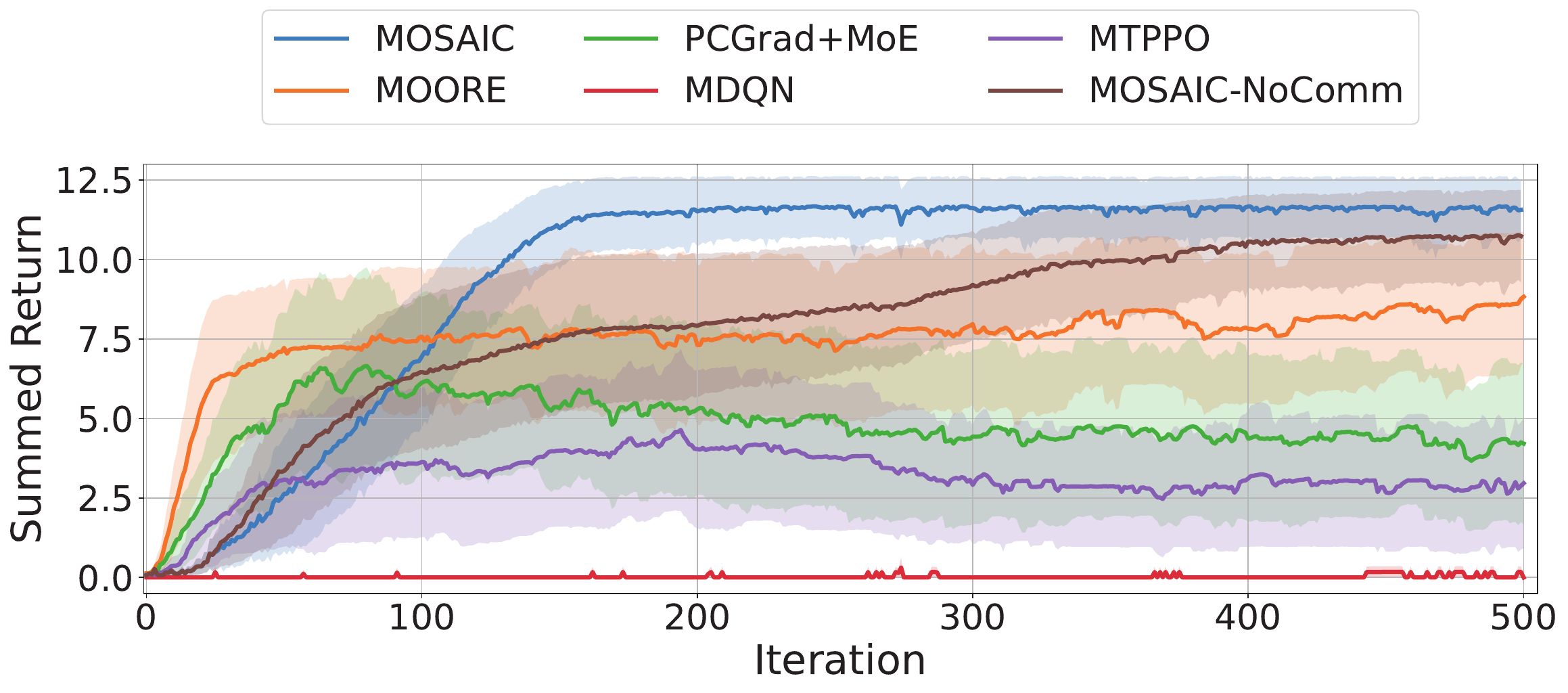}
    \caption{Comparison of MOSAIC to baseline approaches on the 14 task MiniGrid curricula made up of SimpleCrossing and LavaCrossing task variations: average performance for 70 runs with 95\% confidence intervals.}
    \label{fig:minigrid_results}
\end{figure}

The MiniGrid benchmark is a sparse-reward grid-world navigation task with symbolic observations. MOSAIC is evaluated on 14 tasks spanning seven SimpleCrossing and seven LavaCrossing variants, which differ in layout, object placement, and room structure. Total return curves for all methods are shown in Figure~\ref{fig:minigrid_results}, with final and intermediate metrics in Table~\ref{tab:baseline_results}. MOSAIC is compared against the following baselines: Multi-Task PPO (MTPPO) uses a shared backbone without modularity or interference mitigation. Multi-DQN (MDQN) \cite{D'Eramo2020Sharing} attaches task-specific Q-heads to a shared encoder but has no mechanism to avoid interference. PCGrad+MoE \cite{yu2020gradientsurgerymultitasklearning} combines a shared encoder, expert subnetworks, and gradient projection to resolve conflicts. Mixture of Orthogonal Experts (MOORE) \cite{Hendawy2023Multi-TaskExperts} promotes expert diversity via orthogonalization with learned gating to reduce overlap.
MOSAIC achieves the highest final return (11.67), outperforming all baselines. 
MTPPO peaks at 4.64 (30\% of the theoretical maximum), while MDQN fails to learn any tasks, peaking at 0.31. MOORE and PCGrad+MoE improve faster early, reaching 50\% thresholds in 24 (49,152 steps) and 54 iterations (110,592 steps) respectively, but plateau well below MOSAIC and never exceed 75\%. We speculate that centralized models with gradient sharing (MTPPO, PCGrad+MoE) achieve early gains through shared features but suffer interference that limits long-term specialization.
\begin{table}
    \centering
    \small
    \begin{tabularx}{\columnwidth}{lX|XXX}
    \toprule
    Method        & Final perf & 25\% perf & 50\% perf & 75\% perf \\
    \midrule
    MOSAIC (ours)       & \cellcolor{green!10}\textbf{11.67} & 55   & 90   & 119 \\
    MTPPO         & 4.64           & 48   & -    & -   \\
    MDQN          & 0.31           & -    & -    & -   \\
    PCGrad+MoE    & 6.66           & 24   & 54   & -   \\
    MOORE         & 8.83           & \cellcolor{green!10}\textbf{13}   & \cellcolor{green!10}\textbf{24}   & -   \\
    MOSAIC-NoComm & 10.78          & 46   & 87   & 291 \\
    \bottomrule
    \end{tabularx}
    \caption{Performance metrics on the MiniGrid Crossing benchmark. The table reports the final average return and the number of iterations to reach 25\%, 50\% and 75\% of the theoretical maximum. Dashes indicate that the thresholds were not reached during the training.}
    \label{tab:baseline_results}
\end{table}

\subsection{How Similarity Helps Targeted Policy Transfers}
\label{sec:sim_transfer}
\begin{figure}
    \centering
    \includegraphics[width=1.0\columnwidth]{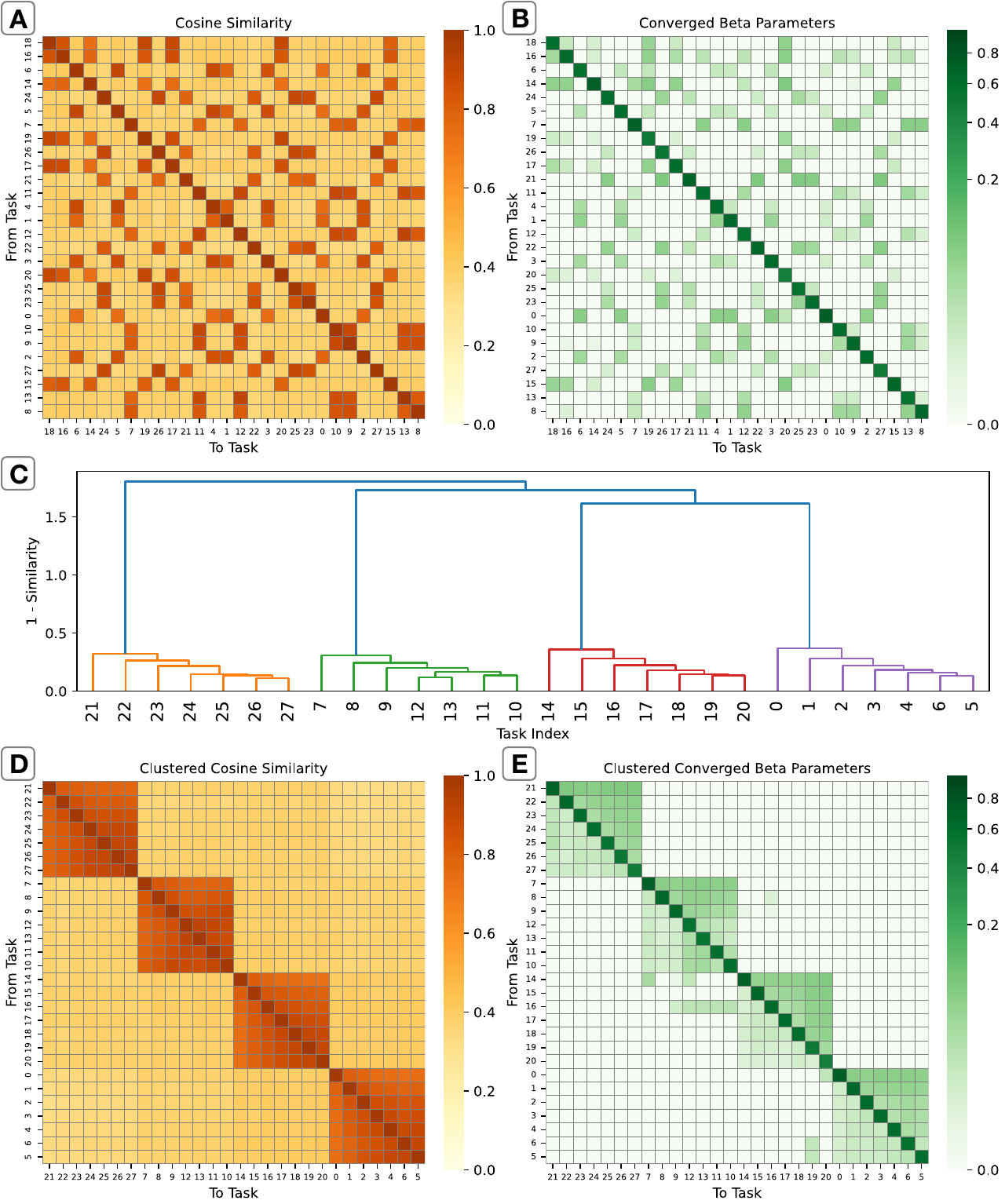}
    \caption{Pairwise cosine similarity and \(\hat{\beta}\) statistics in the image sequence learning benchmark. (A) Cosine similarity matrix. (B) \(\hat{\beta}\) values indicating policy use per task. (C) Cosine similarities clustered using WPGMA, shown as a dendrogram. (D) Cosine similarity matrix reordered by clustering. (E) Clustered \(\hat{\beta}\) values show that similar tasks exhibit the highest policy reuse. Annotated heatmaps of individual clusters are in Appendix~\ref{sec:further_analysis_sim_beta}, Figure~\ref{fig:zoom-grid-heatmaps}.}
    \label{fig:heatmaps}
\end{figure}

Figure~\ref{fig:heatmaps} illustrates the average cosine similarity in the CT-graph benchmark between task embeddings and the converged consolidation parameters \(\hat{\beta}^i_j\) (see Eq.~\eqref{eq:consolidation}),  which provides an indication of the influence of policies derived from task \(j\) when learning task \(i\).

The cosine similarities of tasks presented in random order (Figure~\ref{fig:heatmaps}(A)), when clustered with WPGMA (Figure~\ref{fig:heatmaps}(C)), allow for the reconstruction of the four  groups of tasks, and even the inter-group difficulty relationship. By reordering the matrix using a such clustering, Figure~\ref{fig:heatmaps}(D) illustrates visually the successful reconstruction of task relationships. The  \(\hat{\beta}\) matrices (Figures~\ref{fig:heatmaps}(B) and (E)) confirm that policy reuse and task similarity remain similar even after fine tuning of the policies. Note that, unlike the symmetric similarity matrix, \(\hat{\beta}\) values are asymmetric, reflecting source-to-target policy reuse.

\begin{figure}
    \centering
    \includegraphics[width=1.0\columnwidth]{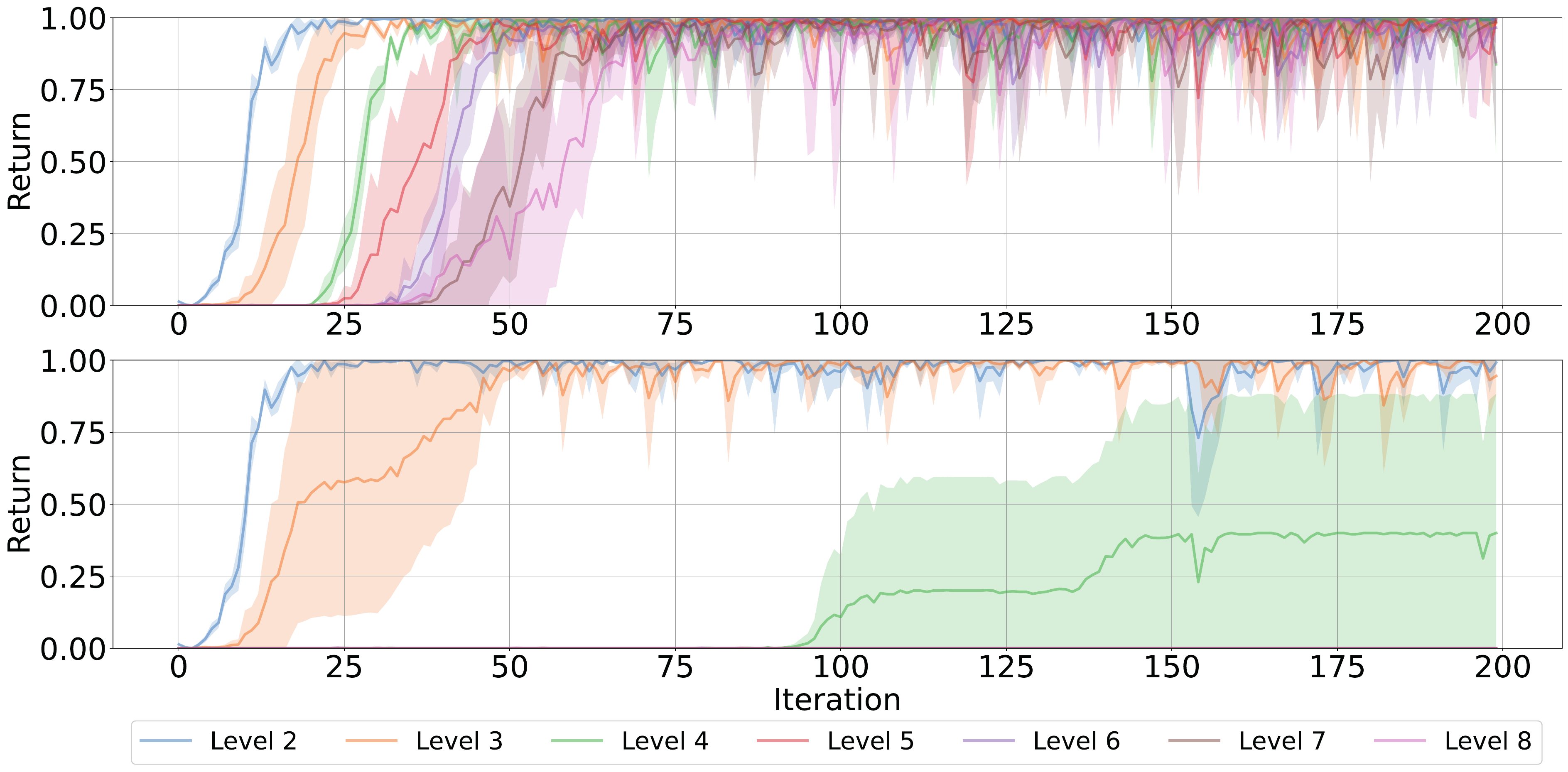}
    \caption{Performance grouped by task complexity, i.e., seven different levels, corresponding to different graph depths, in the image sequence learning problem. Average performance for 20 runs (four tasks and five seeds/task) with
95\% confidence intervals. Communicating MOSAIC agents (top graph) are compared with isolated MOSAIC agents (bottom graph). Communicating agents solve tasks progressively, from the simplest to the hardest tasks. Isolated agents only manage to solve the two simplest tasks, and partially the third, but fail on the four most complex tasks.}
    \label{fig:ctgraph_individual}
\end{figure}

Figure~\ref{fig:ctgraph_individual} presents the performance of agents in the CT-graph grouped by difficulty levels. Interestingly, comparing with isolated learning (bottom graph) suggests that the policies of easier tasks are progressively shared with agents solving harder tasks. This coordination enabling those agents to find successful policies that the same RL algorithm fails to discover when learning in isolation. This analysis helps explain the 170.8\% performance gain of communicating agents over isolated agents (Figure~\ref{fig:ctgraph_minihack_results}(A)): agents tackling harder tasks retrieve, combine, and refine policies from simpler tasks to solve complex problems.

\subsection{Ablation Studies}  
\begin{figure}
    \centering
    \includegraphics[width=1.0\columnwidth]{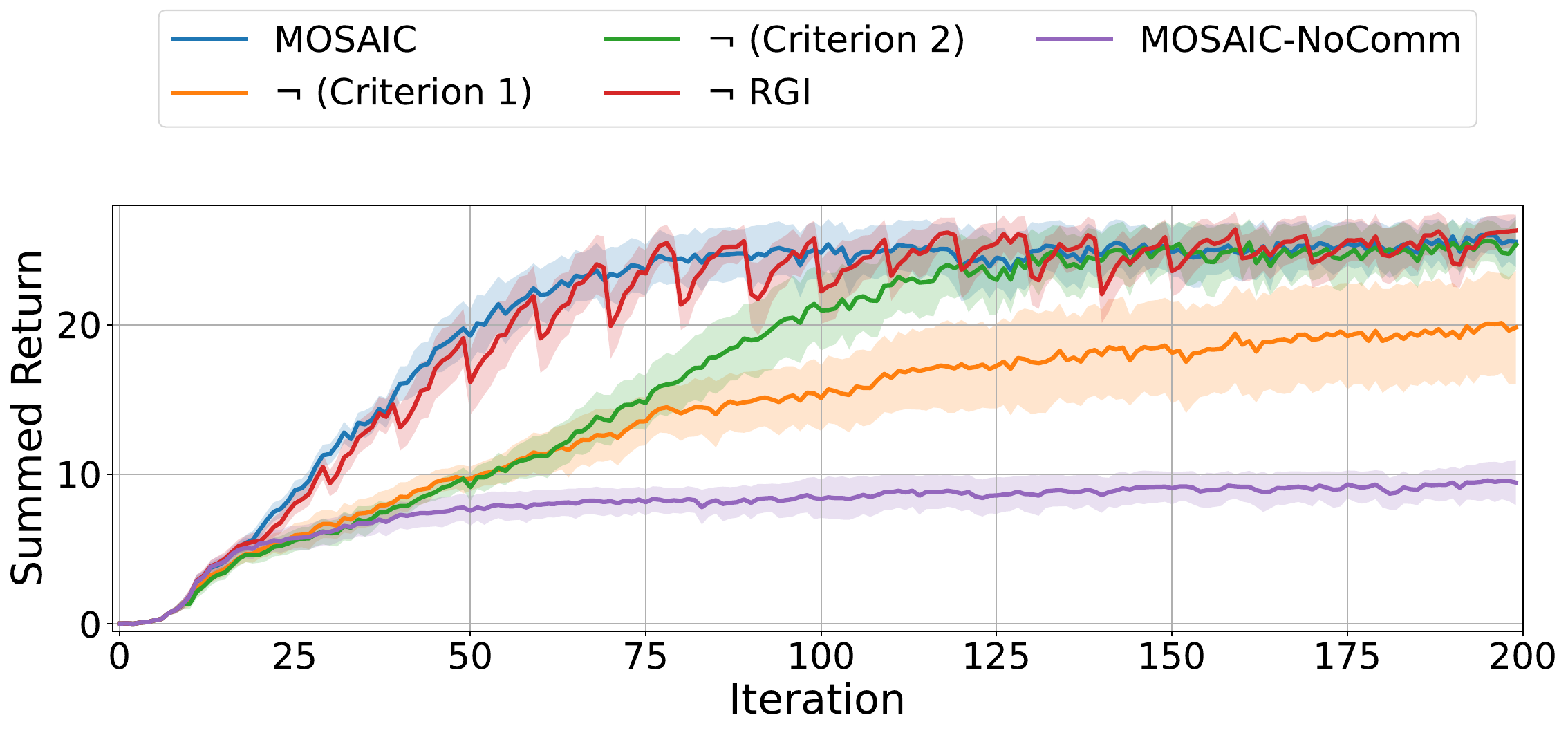}
    \caption{Ablation studies showing average performance over 140 runs (28 tasks, five seeds per task), with 95\% confidence intervals. Variants tested include MOSAIC with  no similarity criterion (\(\neg\) Criterion 1), no reward criterion (\(\neg\) Criterion 2), or no reward-guided initialization (\(\neg\) RGI). MOSAIC-NoComm is included as a baseline. Removing any of these components significantly degrades performance, highlighting their combined importance.}
    \label{fig:main_ablation}
\end{figure}

Ablation studies were conducted to assess the effect on performance of MOSAIC's knowledge selection criteria and reward-guided initialization (RGI) on performance. Figure~\ref{fig:main_ablation} compares three ablated variants of MOSAIC: without cosine similarity-based selection (\(\neg\) Criterion 1), without performance-based selection (\(\neg\) Criterion 2), and  without reward-guided weight initialization (\(\neg\) RGI). MOSAIC-NoComm is included as a reference baseline.

\(\neg\) Criterion 1 reaches a total return of 20.1, whereas MOSAIC reaches 26.0, outperforming it by a factor of 1.29. We speculate that this gap could widen further in settings in which the number of unrelated tasks grows, i.e., where selecting the most relevant knowledge becomes critical. 

\(\neg\) Criterion 2 performs comparable to MOSAIC at the end of training (25.6), but exhibits significantly slower learning. This behavior makes sense because as agents improve their performance, prioritizing higher reward becomes less critical. These two ablations \(\neg\) Criterion 1 and \(\neg\) Criterion 2 reach 50\% of the theoretical maximum (14.0) in 76 (38,912 steps) and 71 iterations (36,352 steps), respectively. MOSAIC reaches this threshold in 37 iterations (18,944 steps), reaching 50\% performance 1.9\(\times\) and 2.0\(\times\) faster, respectively.

\(\neg\) RGI removes reward-guided initialization and instead uses a fixed weighting of 0.5 for the local policy and 0.5 for external policies. This experiment performs comparably with MOSAIC; however, the average return shows periodic instability, characterized by sharp performance drops coinciding with communication events, resembling a sawtooth pattern. This behavior suggests that RGI is crucial to learning stability when integrating external masks.

Further ablation studies are shown in Appendix~\ref{sec:appendx_ablations} to test the impact of query frequency on the performance of MOSAIC, and the impacts of the number of samples and size of the reference distribution on embedding accuracy and performance.

\section{Discussion}
\label{sec:discussion}
The results show that MOSAIC's principles are effective across diverse settings. Agents achieved faster and more successful learning dynamics than isolated agents. In particular, transferring masks from some tasks accelerated learning in other tasks, consistent with findings in centralized lifelong learning \cite{ben2022lifelong}. These gains were most evident in reward-sparse and hierarchical environments, where agents that learned simpler tasks supported those that learned harder ones \cite{Florensa2017AutomaticAgents, Nachum2018Data-EfficientLearning}.

The ablation study highlights the importance of selection based on relevance and utility and of reward-adjusted weighting when integrating new policies. Naïve sharing overwrites stable learning \cite{isele2018selective}, deteriorating performance. These results emphasize the importance of selection when the available tasks and curriculum are unknown \cite{Foerster2016LearningLearning}. The effect of reward-guided initialization suggests that careful policy weighting is important to avoid performance disruption before fine tuning.

It is worth noting that MOSAIC's design takes inspiration from lifelong learning algorithms. Thus, policy composition can easily be extended to previously learned policies of past tasks. Such an extension could deal with continuous streams of evolving tasks \cite{serra2018hat}, supporting scalable and open-ended lifelong learning.

\paragraph{Limitations.}
Although the experiments were conducted in simulation, MOSAIC is designed for decentralized, bandwidth-limited deployments (e.g., robot fleets, on-device perception). Agents exchange only compact embeddings and mask scores while inference stays on-device. Key deployment considerations are cross-site normalization, private/authenticated exchange, and tuning query frequency and size to fit bandwidth.

MOSAIC relies on raw per-iteration reward as a selection and weighting signal, which limits generalization to environments with differing reward functions. Reward-normalized scores or task-progress metrics could improve environment-agnostic reuse. Adaptive or sparse topologies would better mitigate bandwidth scalability constraints \cite{tang2024collaborative}. The policy composition used in MOSAIC is limited to positive combinations only. Other composition or mixture approaches could be tested \cite{He2021TowardsLearning}. Alternative modular skill representations could be used for applications to larger neural networks, e.g., LoRA-like modules and adapters for transformer architectures \cite{hu2022lora, He2021TowardsLearning}. In all such cases, the requirement for a shared backbone or foundation model can reduce the pool of agents able to share knowledge and limit long-term development of progressively more complex skills \cite{pmlr-v97-houlsby19a}.

Communication and selection mechanisms in MOSAIC were introduced as a proof-of-concept to test whether selective transfer is essential for agentic knowledge reuse \cite{DARPA:ShELL, soltoggio2024nature}. Allowing policies to learn when and how to communicate could further increase autonomy \cite{Foerster2016LearningLearning}, but dynamic communication policies raise safety concerns. Agents could learn uncooperative strategies, exploit collective knowledge without contributing, or share harmful policies. Such risks mirror classic game-theoretic challenges and require safeguards. MOSAIC-like systems may also face adversarial attacks exploiting the agent-to-agent nature through model poisoning or policy extraction.

\section{Conclusion}
\label{sec:conclusion}
This work introduced Modular Sharing and Composition in Collective Learning (MOSAIC), where agents search, retrieve, and compose policies obtained from peers learning other tasks. The collective was shown to significantly outperform isolated agents by selecting and fine-tuning task-aligned knowledge. Results show that policy sharing is most effective when implemented selectively, with policy weighting and fine-tuning. Selection also results in the discovery of implicit curricula where simpler tasks help agents learn complex ones faster.

MOSAIC improves interpretability, as composed policies are traceable to their sources and the approach could extend to domains such as language, perception, and control. However, autonomous knowledge sharing carries risks: flawed or misaligned policies may spread rapidly. Future work must ensure that transfers remain safe and aligned with agent objectives. Using selective and modular reuse, MOSAIC can support scalable, adaptive, and collaborative agentic AI in real-world applications.

\section*{Acknowledgements}
This material is based upon work supported by the Defense Advanced Research Projects Agency (DARPA) under contract No. HR001121901 (Shared Experience Lifelong Learning) and the Industrial Robots-as-a-Service (IRaaS) project funded by the EPSRC (EP/V050966/1).

\bibliography{aaai2026}

\clearpage
\appendix
\begin{table}
    \centering
    \small
    \begin{tabularx}{\columnwidth}{lX}
    \toprule
    Symbol        & Definition \\
    \midrule
    \(\Phi\)      & Frozen backbone network\\
    \(\tau\)      & Task \(\tau\)\\
    \(\phi_\tau\) & Randomly initialized learnable mask for task \(\tau\)\\
    \(\phi_k\)    & A mask acquired from a peer\\
    \(\pi_\tau\)  & Policy for task \(\tau\)\\
    \(c\)         & A communication event\\
    \(P_c\)       & A set of masks acquired from peers\\
    \(g(\cdot)\)  & Element-wise binarization function\\
    \(\ell\)      & Layer \(\ell\) of the network \\
    \(\gamma^*\)  & Optimal transport problem \\
    \(v_\tau\)    & Wasserstein task embedding for task \(\tau\)\\
    \(\mu_\tau\)  & Empirical distribution for task \(\tau\)\\
    \(\mu_0\)     & Fixed synthetic reference distribution \\
    \(\beta\)     & Learnable mixture weights for linear combination \\
    \(\theta\)    & Cosine similarity threshold\\
    \(\bar{r}\)   & Normalized iteration reward scalar\\
    \(N\)         & Number of SAR samples for embedding computation\\
    \(M\)         & Number of reference points in the synthetic distribution\\
    \(\mathbb{I}_\mathrm{align}\) & Task similarity selection heuristic\\
    \(\mathbb{I}_\mathrm{perf}\)  & Performance selection heuristic\\
    
    \bottomrule
    \end{tabularx}
    \caption{Table of important notations in the paper.}
    \label{tab:notation}
\end{table}
\section{Significance Testing}
\label{sec:significance}
Deep reinforcement learning experiments exhibit substantial variability across random seeds and environment stochasticity \cite{henderson2018deep}. To improve the robustness and interpretability of the results, we follow the recommendations of \citep{colas2018many} and report both confidence intervals and formal significance tests on the results.

For each benchmark, each algorithm is evaluated on multiple agents. For every agent, 5 independent algorithm seeds and 5 environment seeds. A difference test is computed using the Welch t-test and bootstrap confidence interval (BCI). Table~\ref{tbl:sig-test} reports the results of this test for the results of each benchmark.
\begin{table*}[h]
    \centering
    \begin{tabularx}{\textwidth}{l|llll}
        \toprule
        \multicolumn{5}{c}{\textbf{CT-graph ISL}}\\
        \midrule
        Method        & Final Perf (mean \(\pm\) 95\% CI) & \(\Delta\) (MOSAIC \(-\) Method) & Welch \(p\) (vs MOSAIC) & 95\% BCI of \(\Delta\)\\
        \midrule
        MOSAIC        & 0.91 [0.87, 0.95]              & -      & -          & -                \\
        \rowcolor{green!6}
        MOSAIC-NoComm & 0.34 [0.26, 0.42]              & 0.58   & 1.74e-28   & [0.49, 0.66]     \\
        \toprule
        \multicolumn{5}{c}{\textbf{MiniGrid Crossing}}\\
        \midrule
        Method        & Final Perf (mean $\pm$ 95\% CI) & $\Delta$ (MOSAIC $-$ Method) & Welch $p$ (vs MOSAIC) & 95\% BCI of $\Delta$\\
        \midrule
        MOSAIC        & 0.83 [0.75, 0.90]              & -      & -          & -                \\
        \rowcolor{red!6}
        MOSAIC-NoComm & 0.77 [0.68, 0.86]              & 0.06   & 3.26e-01   & [-0.06, 0.18]    \\
        \rowcolor{green!6}
        MOORE         & 0.63 [0.52, 0.74]              & 0.20   & 3.98e-03   & [0.07, 0.32]     \\
        \rowcolor{green!6}
        PCGrad+MoE    & 0.30 [0.20, 0.41]              & 0.53   & 6.37e-13   & [0.39, 0.65]     \\
        \rowcolor{green!6}
        MDQN          & 0.00 [0.00, 0.00]              & 0.83   & 3.82e-32   & [0.75, 0.90]     \\
        \rowcolor{green!6}
        MTPPO         & 0.21 [0.12, 0.31]              & 0.61   & 4.40e-18   & [0.49, 0.73]     \\
        \toprule
        \multicolumn{5}{c}{\textbf{MiniHack MultiRoom}}\\
        \midrule
        Method        & Final Perf (mean \(\pm\) 95\% CI) & \(\Delta\) (MOSAIC \(-\) Method) & Welch \(p\) (vs MOSAIC) & 95\% BCI of \(\Delta\)\\
        \midrule
        MOSAIC        & 0.64 [0.54, 0.74]              & -      & -          & -                \\
        \rowcolor{green!6}
        MOSAIC-NoComm & 0.27 [0.15, 0.38]              & 0.37   & 2.06e-06   & [0.22, 0.51]     \\
        \bottomrule
    \end{tabularx}
    \caption{Total evaluation performance and significance testing CT-graph, MiniGrid and MiniHack benchmarks. \(\Delta\) denotes MOSAIC \(-\) Method. Rows highlighted in green indicate methods for which MOSAIC is significantly better (Welch's \(t\)-test \(p<0.05\) and 95\% bootstrap CI of \(\Delta\) excludes 0).}
    \label{tbl:sig-test}
\end{table*}

\section{Extended Related Work} \label{sec:extended_related_work} This section provides additional information on the work related to MOSAIC and clarifies its unique contributions within the broader landscape.

\paragraph{Modular Representations and Masking.} Modular policies promote parameter reuse by isolating task-specific substructures. Binary masks define sparse sub-regions of shared networks, reducing interference and enabling structured composition \cite{mallya2018piggyback, mallya2018packnet, ramanujan2020s, serra2018hat, wortsman2020supermasks, koster2022signing_sm}. Some approaches compose masks to support task transfer \cite{Ostapenko2021ContinualComposition, mendez2022neuralcomposition, GummadiSHELS:BOUNDARIES, ben2022lifelong}. These methods typically require predefined task boundaries or centralized coordination. MOSAIC differs by enabling decentralized, performance-driven mask selection without explicit task labels.

Mixture-of-Experts (MoE) architectures route inputs to sparse expert modules via learned gates \cite{shazeer2017outrageouslylargeneuralnetworks, fedus22switchtransformers}, with variants adapted to multi-task and lifelong settings \cite{gururangan-etal-2020-dont, mendez2021lifelong}. MOORE imposes orthogonality constraints across experts to promote representational diversity \cite{Hendawy2023Multi-TaskExperts}. Related work in modular meta-learning emphasizes structural reuse through task-level module composition \cite{pmlr-v87-alet18a, mittal2022compositionplan}. These approaches require training routing mechanisms or meta-learning procedures. MOSAIC instead uses Wasserstein-based similarity to automatically select relevant modules without additional training overhead.

LoRA adds task-specific low-rank adapters to fixed models for efficient specialization \cite{hu2022lora}. Neural Module Networks support compositional inference via dynamically assembled submodules \cite{andreas16neural, pmlr-v139-araki21a}. MultiCriticAL isolates value functions per task to mitigate interference in actor-critic training \cite{MysoreMULTI-CRITICSTYLE}. Recent robotic lifelong learning systems adopt modular structures with Bayesian non-parametric inference to preserve and combine task-specific knowledge \cite{meng2025preserving}. Modular compositionality on policies enables agents to dynamically combine pre-trained skills to solve new tasks in a scalable and interpretable manner \cite{10645586}. MOSAIC extends these ideas by combining binary mask modularity with performance-based selection and similarity-driven policy integration in a fully decentralized manner.

\paragraph{Task Similarity and Inference.} Task inference enables modular adaptation without task labels. Latent and Bayesian models estimate embeddings from trajectories \cite{Zintgraf2019VariBAD:Meta-Learning, Achille_2019_ICCV}. Non-parametric and prototype-based methods define task similarity via behavioral statistics \cite{yarats2021reinforcementlearningprototypicalrepresentations, Chu2023Meta-ReinforcementClustering, liu2022wasserstein, Kolouri2020WassersteinLearning, dick2024statistical}. Taskonomy \cite{zamir2018taskonomydisentanglingtasktransfer} and Task2Vec \cite{Achille_2019_ICCV} provide task embeddings for transfer learning. Recent work explores which tasks benefit from joint learning \cite{pmlr-v119-standley20a}. These methods typically require centralized computation or predefined task structures. MOSAIC computes Wasserstein-based similarity directly from agent trajectories in a decentralized fashion, enabling online task relationship discovery.

Wasserstein barycenter methods enable structure-aware ensembling by incorporating semantic side information through optimal transport distances \cite{Dognin2019WassersteinEnsembling}. Learning embeddings into entropic Wasserstein spaces offers a flexible alternative to Euclidean embeddings for capturing rich relational structures \cite{Frogner2019LearningSpaces}. MOSAIC leverages these optimal transport principles specifically for RL trajectory comparison and module selection.

Contrastive objectives such as CURL \cite{Srinivas2020CURL:Learning} and ATC \cite{Stooke2020DecouplingLearning} improve representation learning, though they typically require encoder training and large batch sizes. Structure-aware distillation has also been explored in incremental graph-based settings \cite{Wang2021GraphSystems}, offering another avenue for modular task inference. MOSAIC avoids the need for contrastive training by directly computing trajectory similarities for module selection.

Gradient-based meta-RL methods such as MAML \cite{finn2017model} optimize for rapid task adaptation through fine-tuning, while variational approaches like VariBAD \cite{Zintgraf2019VariBAD:Meta-Learning} condition policy learning on inferred latent variables. In contrast, non-parametric methods that reuse task-conditional modules based on similarity measures offer an alternative path to fast generalization without the need for gradient-based adaptation. MOSAIC follows this non-parametric approach.

\paragraph{Model Merging and Upcycling.} Emerging techniques enable the combination and repurposing of existing models through weight-space merging or routing. \citet{Wortsman2022ModelTime} and \citet{Rame2023RewardedRewards} propose interpolating parameters from multiple fine-tuned models to improve generalization. TIES-Merging \cite{Yadav2023TIES-Merging:Models} resolves interference between merged models. \citet{Komatsuzaki2022SparseCheckpoints} and \citet{Sukhbaatar2024Branch-Train-MiX:LLM} demonstrate sparse upcycling: extracting sparse modular experts from dense checkpoints. These strategies provide alternatives to training from scratch and are critical for decentralized reuse. MOSAIC extends these ideas to multi-agent RL by combining similarity-driven linear policy combinations with binary mask modularity.

\paragraph{Adaptive and Elastic Architectures.} Recent work explores modular and resource-aware architectures. MatFormer \cite{Devvrit2023MatFormer:Inference} and Mixture-of-Depths \cite{Raposo2024Mixture-of-Depths:Models} dynamically allocate compute and adjust routing depths in transformer models. These approaches support efficient inference and specialization while preserving modular structure, principles aligned with our use of task-conditional module composition. MOSAIC applies similar principles to multi-agent systems through selective module reuse based on performance signals.

\paragraph{Hierarchical and Curriculum Learning.} Hierarchical RL tackles long-horizon credit assignment and sparse reward signals. HAC \cite{Levy2017LearningHindsight} introduces layered actor-critic policies operating at different temporal resolutions. CAMRL \cite{Huang2023Curriculum-BasedLearning} proposes curriculum-driven asymmetric MTRL between teacher and student agents. Our method complements these paradigms by achieving curriculum-like generalization via unsupervised trajectory similarity and decentralized reuse without explicit hierarchical structures.


\paragraph{Selective Transfer Learning.} Selective transfer methods determine what and where to transfer knowledge. \citet{Chu_2013_CVPR} introduces selective transfer machines for personalized learning. \citet{jang2019learningtransfer} learns what and where to transfer in deep networks. SpotTune \cite{guo2018spottunetransferlearningadaptive} enables adaptive fine-tuning through selective transfer. These methods typically operate in supervised settings with labeled tasks. MOSAIC extends selective transfer to decentralized multi-agent RL through Wasserstein-based similarity and performance signals.

\section{Extending MOSAIC to Lifelong Reinforcement Learning}
\label{sec:mask_composition_ll}

MOSAIC can be naturally extended to a lifelong learning (LL) setting \cite{delangecl2022survey,kudithipudi2022biological}, in which each agent learns a sequence of tasks while minimizing catastrophic forgetting and promoting forward transfer. LL MOSAIC agents can exploit their knowledge integration capabilities to accumulate knowledge that may derive from other agents, or from their own learning on previously encountered tasks. MOSAIC employs knowledge composition approaches that were developed in the field of lifelong reinforcement learning \cite{ben2022lifelong}.

In the experimental settings for MOSAIC, however, we employed a one task per agent approach to maintain the focus on knowledge sharing and composition. This setting can be extended to a more general case in which one agent is not limited to learning one task. Therefore, MOSAIC can be easily adapted to a fully-fledged lifelong learning scenario by modifying the linear combination and \(\beta\) parameter adjustment.

Let agent \(i\) currently be learning task \(\tau_k+1\). Its mask is composed from three sources: the current task mask \(\phi_{\tau_k+1}\), prior task masks \(\{\phi_{\tau_k}\}_{k=1}^{K}\), and peer masks \(\{\phi_{j}\}_{j=1}^{n}\), acquired from other agents. The composed mask for task \(\tau_{k+1}\) for the forward pass is thus given by,
\begin{equation}
    \phi_{\tau_{k+1}}^\text{lc} = \beta_{\tau_{k+1}} \phi_{\tau_{k+1}} + \sum_{k=1}^k \beta_{\tau_k}  \phi_{\tau_k} + \sum_{j=1}^n \beta_j \phi_j,
\end{equation}
where each \(\beta\) is a scalar weight normalized via softmax. In this case, the initialization problem includes setting the weighting for previous task masks, the current task mask, and the newly acquired masks from other agents. One simple approach could simply be to extend Eq.~\eqref{eq:reward-guided-init} to 
\begin{equation}
    \beta_{\tau_{k+1}} = \frac{1}{3} + \frac{2}{3}r, \quad \beta_{\tau_k} = \frac{(1 - r)}{3k}, \quad \beta_j = \frac{(1 - r)}{3n}.
\end{equation}
The addition of lifelong learning presents an additional challenge. Boundaries in the sequence of tasks learned by an agent need to be identified, so that a new mask can be learned for each new task. One solution to this could be to use the cosine similarity computed from the Wasserstein embeddings to monitor the incoming data stream and decide when a new task starts using statistical methods \cite{dick2020detecting,dick2024statistical}.

\section{Additional Ablation Studies}
\label{sec:appendx_ablations}
\begin{figure}
    \centering
    \includegraphics[width=1.0\columnwidth]{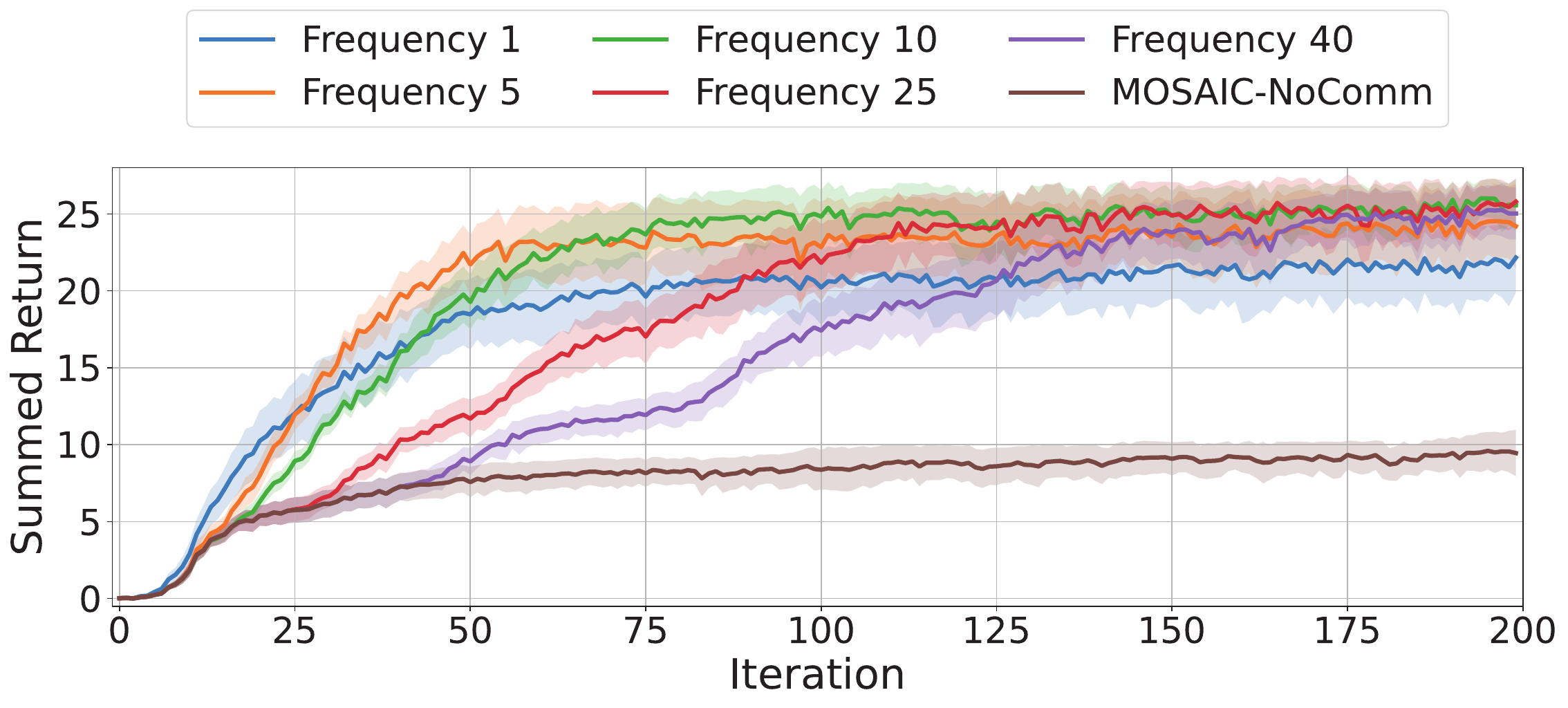}
    \caption{Comparison of MOSAIC with varying communication frequencies highlights the effect on performance in the image sequence learning problem. Striking the right balance in communication frequency is critical: infrequent communication can lead to missed performance gains, while communicating at every iteration may not allow sufficient time for effective local adaptation.}
    \label{fig:freq_ablation}
\end{figure}

\paragraph{Comparison of Sharing Frequencies.} An ablation study was conducted to assess the effects of communication frequency on MOSAIC's performance in the image sequence learning benchmark. Specifically, comparing frequencies of 1, 5, 10 (default), 25, and 40 iterations, where the frequency indicates how often agents search, retrieve and compose peer masks. As shown in Figure~\ref{fig:freq_ablation}, excessive communication (frequency of 1) results in performance degradation. In this setting, newly acquired masks continually override the agent's own beta parameter adjustments, preventing meaningful adaptation and fine-tuning. Frequencies of 5 and 10 yield comparable early performance, with a frequency of 10 achieving higher final return. Conversely, with infrequent communication (25 or 40 iterations) agents have fewer opportunities for useful knowledge transfer, leading to weaker overall performance gains. These results highlight the importance of balancing between querying for peer knowledge often enough for a benefit to occur, and allowing agents to then tailor peer masks towards their own tasks.

\begin{figure}
    \centering
    \includegraphics[width=1.0\columnwidth]{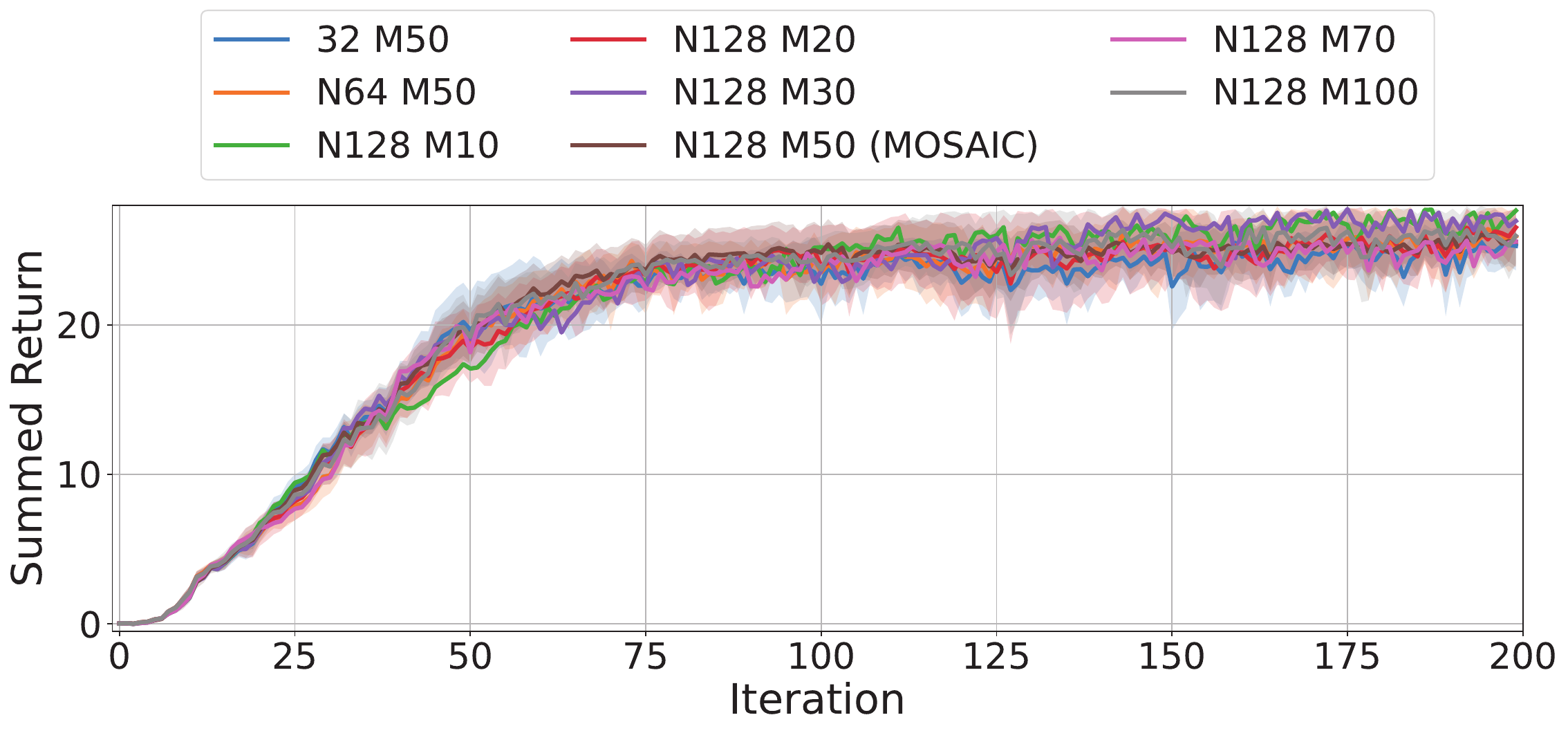}
    \caption{Comparison of MOSAIC with varying communication frequencies highlights the effect on performance in the image sequence learning problem. Striking the right balance in communication frequency is critical: infrequent communication can lead to missed performance gains, while communicating at every iteration may not allow sufficient time for effective local adaptation.}
    \label{fig:detect_ablation}
\end{figure}

\paragraph{Comparing Effects of Number of Samples and Reference Size on Embedding Similarity.} We conducted a grid search on the number of samples per embedding \(N\) (Eq.~\eqref{eq:embedding_distribution}) and the size of the reference distribution \(M\) (Eq.~\eqref{eq:reference_distribution}) to determine their effects on similarity computations and performance on the image sequence learning benchmark. Results shown in Figure~\ref{fig:detect_ablation} indicate no statistically significant changes from using different parameters of \(N\) and \(M\), on the resulting performance. 

\begin{figure*}
    \centering
    \includegraphics[width=1.0\textwidth]{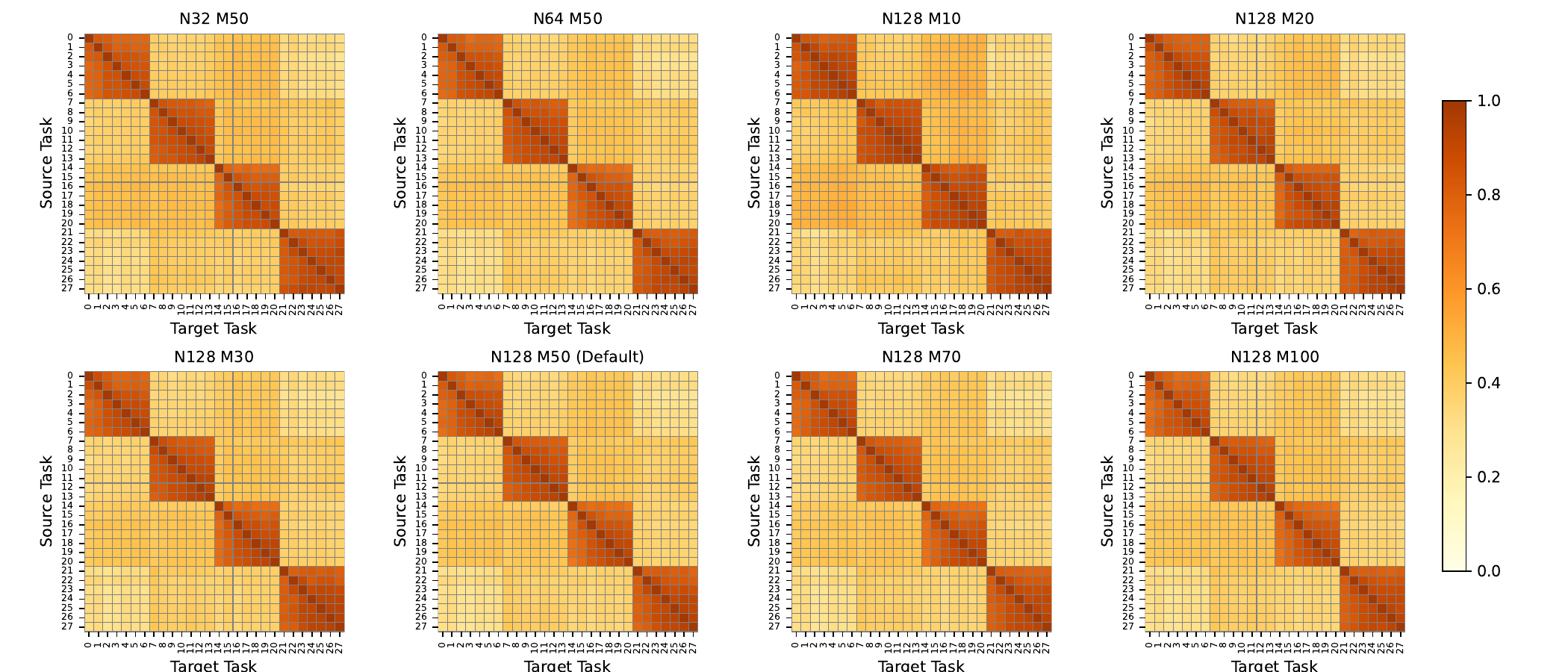}
    \caption{Cosine similarity heatmaps for configurations of \(N\) and \(M\). Cross-cluster cosine similarity increases with smaller values of \(N\) and \(M\), leading to instances of cross-cluster sharing. Higher values increase the accuracy of embedding cosine similarity.}
    \label{fig:detect_ablation_similarity}
\end{figure*}

\begin{figure*}
    \centering
    \includegraphics[width=1.0\textwidth]{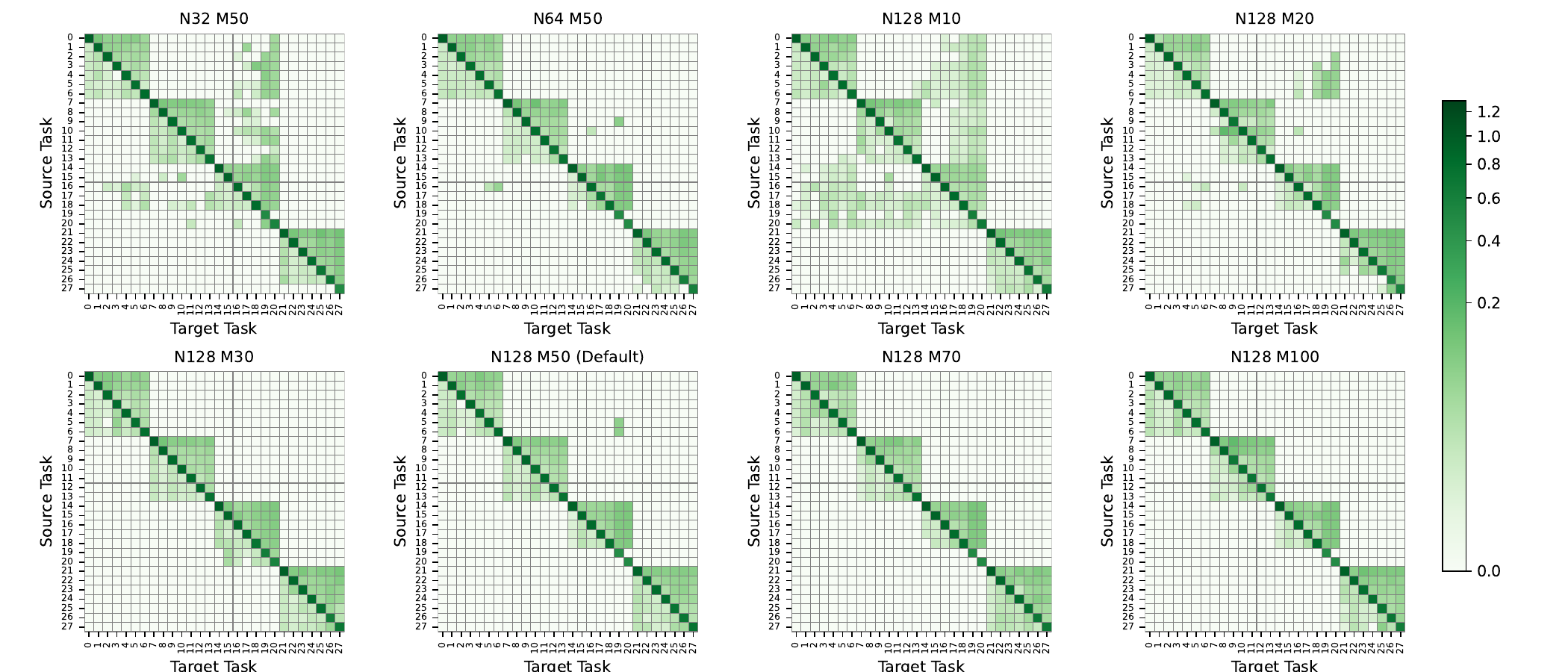}
    \caption{Beta parameters heatmaps for configurations of \(N\) and \(M\). Less accurate cosine similarity estimates result in the transfer of masks. This then leads to the reuse of these masks in the linear combination.}
    \label{fig:detect_ablation_beta}
\end{figure*}

However, Figures~\ref{fig:detect_ablation_similarity} and \ref{fig:detect_ablation_beta} indicate that using smaller values of \(N\) and \(M\) leads to increased knowledge reuse across task distributions. In more challenging settings, this may lead to catastrophic interference where misleading policies result in performance degradation. The effects of this are reduced in MOSAIC through the finetuning process in the linear combination of masks.

Table~\ref{tbl:cross_dist_sharing_freq} reports the proportion of sharing within vs. cross-task distribution, for average cosine similarity and average beta parameter values. Within-cluster cosine similarity remains high across all configurations (0.84–0.88), indicating robust internal coherence among task groups. The cosine similarity between task distributions is more variable (0.38–0.43), with higher values observed at lower \(M\), suggesting decreased separability between groups when mask capacity is restricted. Beta parameters exhibit a strong asymmetry: cross-distribution reuse is negligible (\textasciitilde
0.000–0.003), while reuse within each cluster remains stable (\textasciitilde
0.036–0.041). These trends confirm that agents primarily leverage intra-cluster task knowledge and that both sampling and masking hyperparameters influence the clarity and utility of the learned compositional structure.
\begin{table*}[ht]
    \centering
    \small
    \begin{tabularx}{\textwidth}{lcccccccc}
    \toprule
    \textbf{Metric} & \textbf{N32 M50} & \textbf{N64 M50} & \textbf{N128 M10} & \textbf{N128 M20} & 
    \textbf{N128 M30} & \textbf{N128 M50} & \textbf{N128 M70} & \textbf{N128 M100} \\
    \midrule
    Cross-Similarity     & 0.3887 & 0.3895 & 0.4288 & 0.4135 & 0.3993 & 0.4047 & 0.4054 & 0.3956 \\
    Within-Similarity    & 0.8779 & 0.8677 & 0.8425 & 0.8482 & 0.8559 & 0.8529 & 0.8576 & 0.8571 \\
    Cross-Beta           & 0.0021 & 0.0006 & 0.0029 & 0.0019 & 0.0013 & 0.0006 & 0.0006 & 0.0000 \\
    Within-Beta          & 0.0384 & 0.0368 & 0.0395 & 0.0377 & 0.0374 & 0.0386 & 0.0397 & 0.0411 \\
    \bottomrule
    \end{tabularx}
    \caption{The average cross-cluster and within-cluster cosine similarity and beta parameter values across various \(N, M\) configurations. Average cosine similarity within clusters decreases as \(N\) and \(M\) increase. The average beta parameter values also increase as \(N\) and \(M\) increase, suggesting increased transfer between tasks driven by more accurate cosine similarity.}
    \label{tbl:cross_dist_sharing_freq}
\end{table*}

\section{Additional Experiment Analysis}
\label{sec:appendix_experiments}
\begin{figure}
    \centering
    \includegraphics[width=1.0\columnwidth]{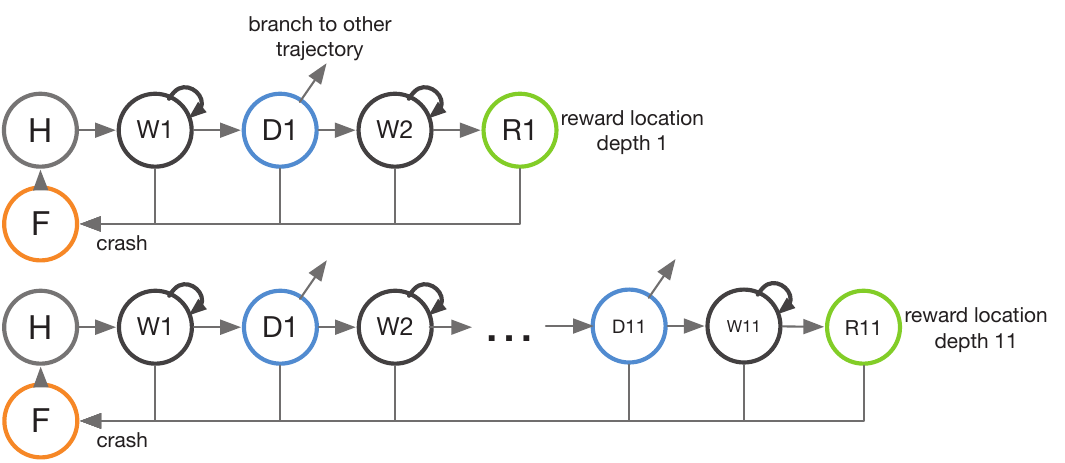}
    \caption{Illustration of the image sequence learning problem. Each graph is comprised of various states, represented by procedurally generated images. The agent must learn to navigate these image states and acquire a sparse reward. (Top) shows an example of a depth-1 graph. Bottom shows a depth 11 graph along the same trajectory.}
    \label{fig:ctgraph}
\end{figure}

Figure~\ref{fig:ctgraph} illustrates an example of the sequence-learning problem based on \cite{soltoggio2019ctgraph}. The agent must navigate a series of nodes labeled as H (home), W (wait state), D (decision state), and R (reward state). For the experiments in this paper, the wait state probability is set to zero, resulting in reduced complexity.

\begin{figure}
    \centering
    \includegraphics[width=0.10\textwidth]{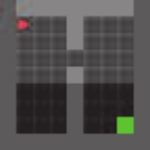}
    \includegraphics[width=0.10\textwidth]{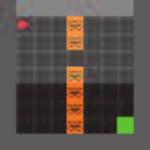}
    \includegraphics[width=0.10\textwidth]{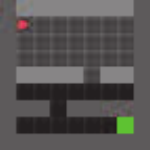}
    \includegraphics[width=0.10\textwidth]{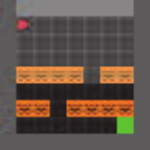}
    \includegraphics[width=0.10\textwidth]{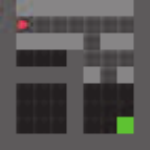}
    \includegraphics[width=0.10\textwidth]{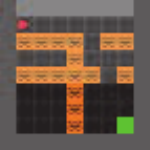}
    \includegraphics[width=0.10\textwidth]{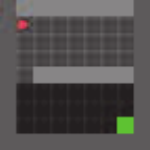}
    \includegraphics[width=0.10\textwidth]{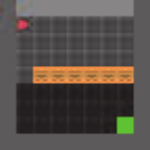}
    \includegraphics[width=0.10\textwidth]{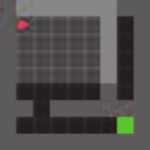}
    \includegraphics[width=0.10\textwidth]{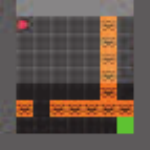}
    \includegraphics[width=0.10\textwidth]{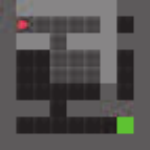}
    \includegraphics[width=0.10\textwidth]{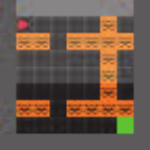}
    \includegraphics[width=0.10\textwidth]{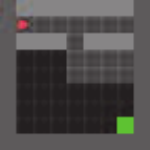}
    \includegraphics[width=0.10\textwidth]{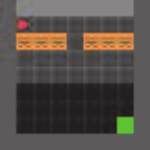}
    \caption{Illustrations of the MiniGrid SimpleCrossing and LavaCrossing curriculum used in our experiments. Each SimpleCrossing variation is accompanied by a LavaCrossing counterpart. This approach tests whether MOSAIC agents can transfer knowledge across different observation spaces while maintaining agent specialization.}
    \label{fig:minigrid}
\end{figure}

Figure~\ref{fig:minigrid} shows examples of the MiniGrid curriculum used in our MiniGrid Crossing experiments. Agents are expected to learn useful behaviors/trajectories to navigate around obstacles in the form of walls or lava, and reach the goal state (green). This setup tests to see if agents can identify similarities in the structure of the tasks, irrespective of the changing observations (wall, lava), and transfer useful policies from SimpleCrossing to LavaCrossing variations, as well as on other SimpleCrossing/LavaCrossing variations.

\begin{figure}
    \centering
    \includegraphics[width=1.0\columnwidth]{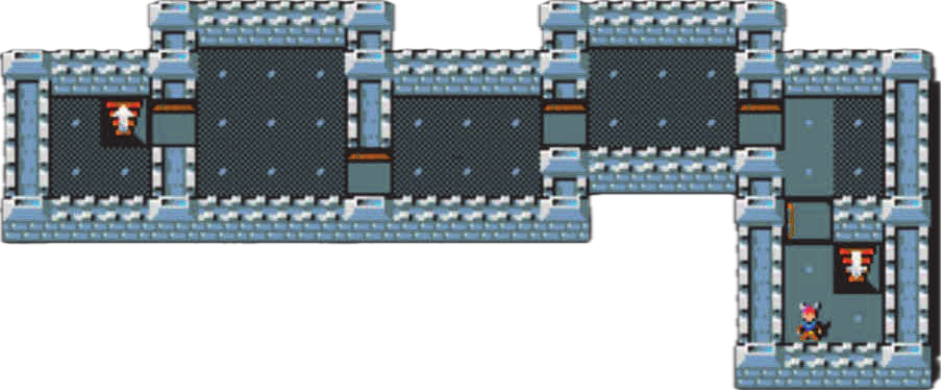}
    \caption{Example illustration of the MiniHack MultiRoom benchmark. Each environment consists of multiple rooms of varying sizes, connected by closed (but unlocked) doors. Tiles with upwards arrows indicate starting locations, and tiles with downwards arrows indicate goal locations.}
    \label{fig:minihack}
\end{figure}

Figure~\ref{fig:minihack} shows an example of a MiniHack MultiRoom task. The start and goal states are represented by squares with up and down arrows, respectively. The agent is expected to navigate through each room, learning to navigate towards and open doors along the way until it reaches the goal state in the final room.

\begin{figure*}
    \centering
    \includegraphics[width=1.0\textwidth]{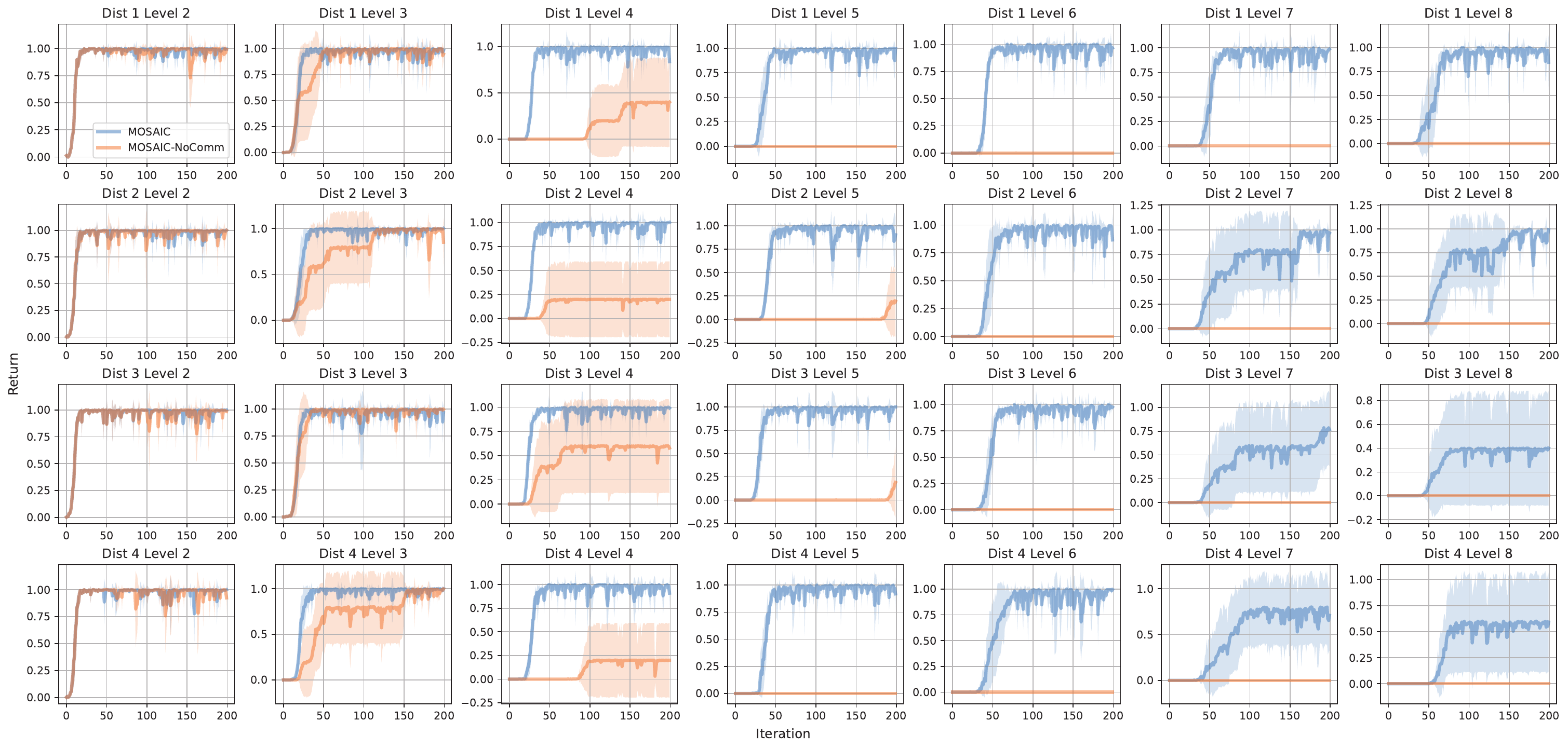}
    \caption{Individual task performance on the ISL benchmark.}
    \label{fig:mctgraph_individual_expanded}
\end{figure*}

\begin{figure*}
    \centering
    \includegraphics[width=1.0\textwidth]{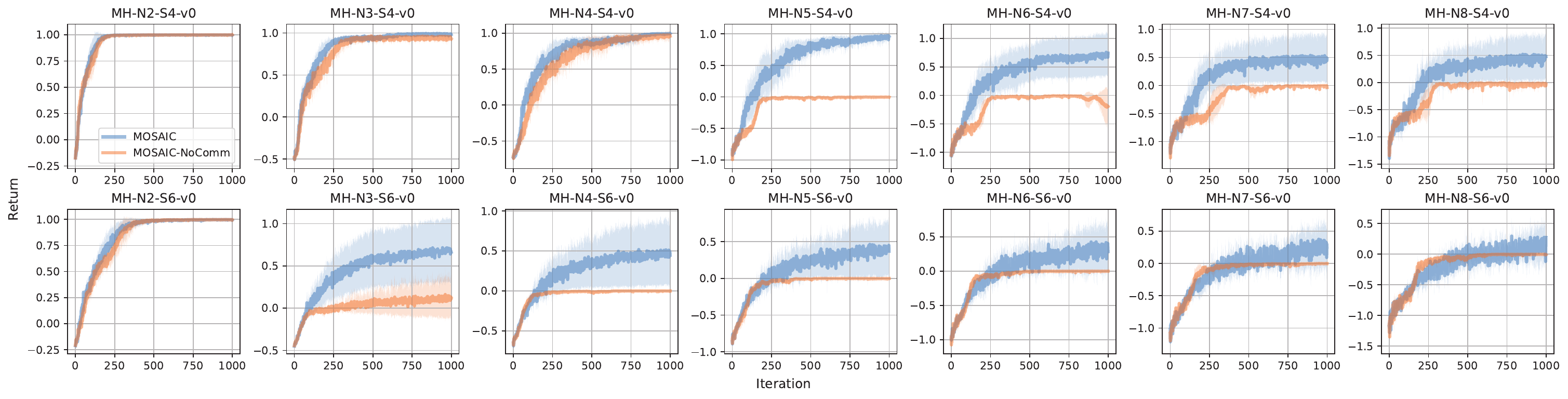}
    \caption{Individual task performance on the MiniHack benchmark.}
    \label{fig:minihack_individual}
\end{figure*}

\begin{figure*}
    \centering
    \includegraphics[width=1.0\textwidth]{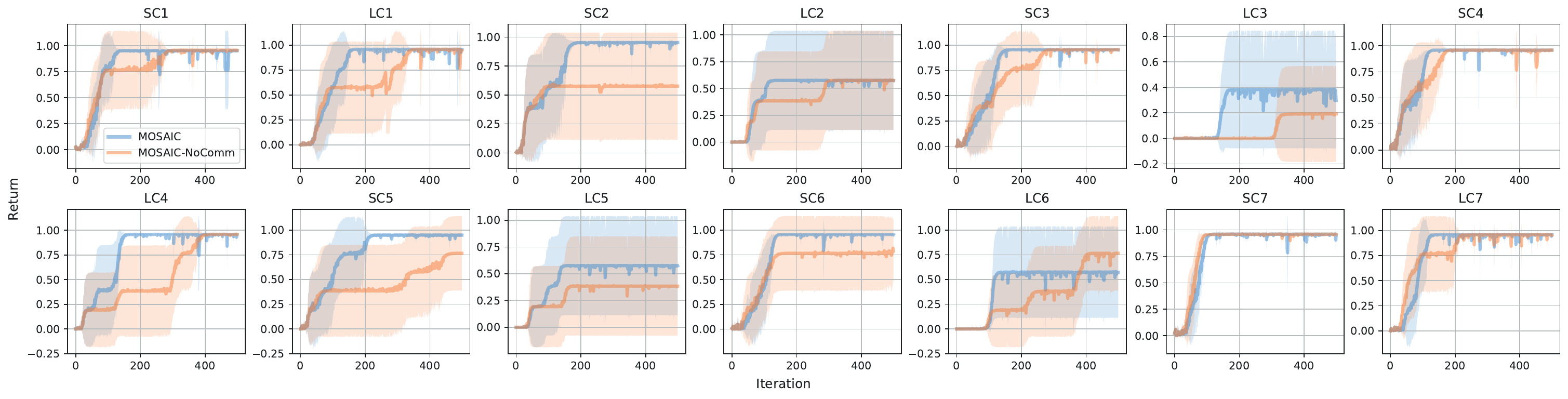}
    \caption{Individual task performance on the MiniGrid benchmark.}
    \label{fig:minigrid_individual}
\end{figure*}

Figures~\ref{fig:mctgraph_individual_expanded}, \ref{fig:minihack_individual}, and \ref{fig:minigrid_individual} show the individual task performance curves for the image sequence learning setting, MiniHack MultiRoom, and MiniGrid Crossing, respectively.

\subsection{Further Analysis of Similarity and Converged Beta Parameters}
\label{sec:further_analysis_sim_beta}
\begin{figure*}[t]
    \centering
    \includegraphics[width=1.0\textwidth]{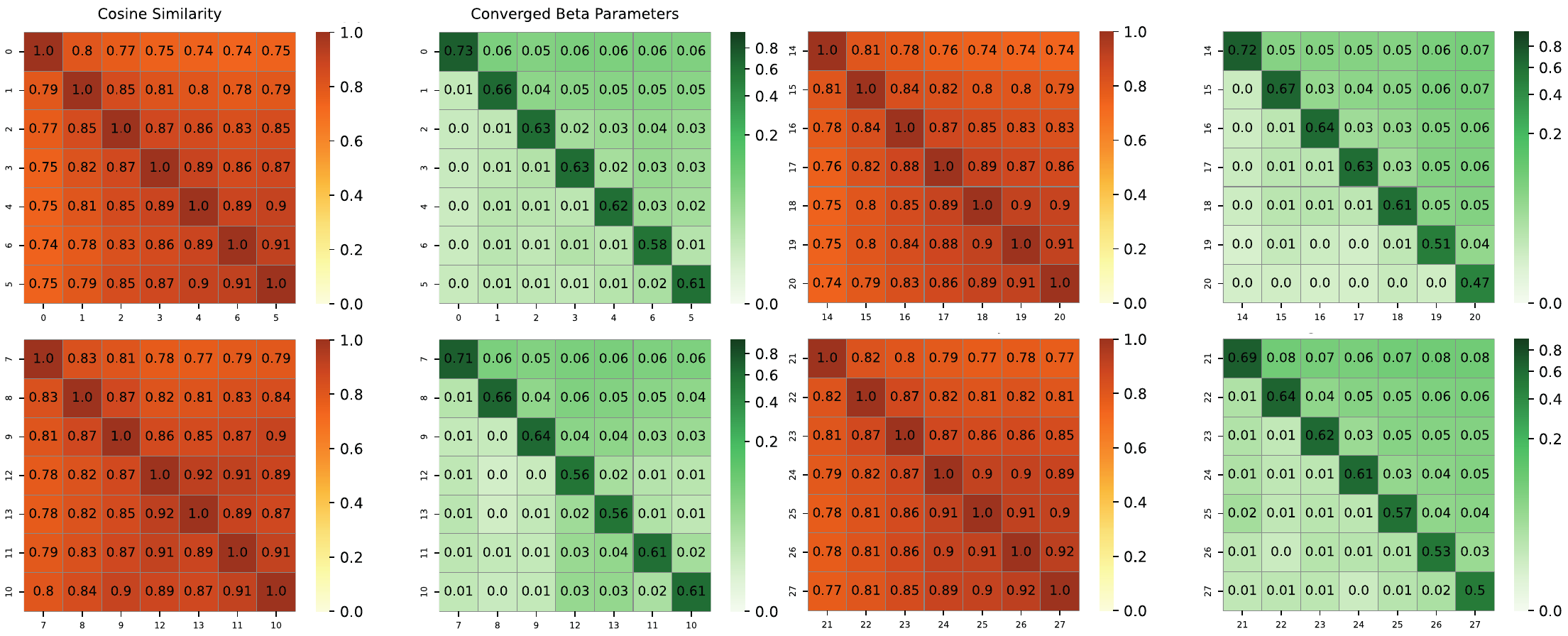}
    \caption{Individual heatmaps of each task distribution cluster from the image sequence learning results shown in Figure~\ref{fig:heatmaps}. Cosine similarity is highest between each task and the next task. Beta parameters indicate that cosine similarity has an inversely proportional relationship with cosine similarity, with foundational prior tasks having more weight. E.g., task 0 has the most weight for reuse in task 5.}
    \label{fig:zoom-grid-heatmaps}
\end{figure*}

The image sequence learning benchmark consists of four distinct image datasets, resulting in four independent state spaces. Four unique trajectories are chosen, one for each image dataset. Seven hierarchically related tasks are selected from each trajectory-dataset combination resulting in a total of 28 tasks with four clusters denoted by (0 to 6), (7 to 13), (14 to 20), (21 to 27). Figure~\ref{fig:zoom-grid-heatmaps} shows the annotated results of the heatmaps presented in Figure~\ref{fig:heatmaps}(D) and (E) for each of these clusters, identified by the WPGMA (Figure~\ref{fig:heatmaps}(C)).

\begin{figure}
    \centering
    \includegraphics[width=1.0\columnwidth]{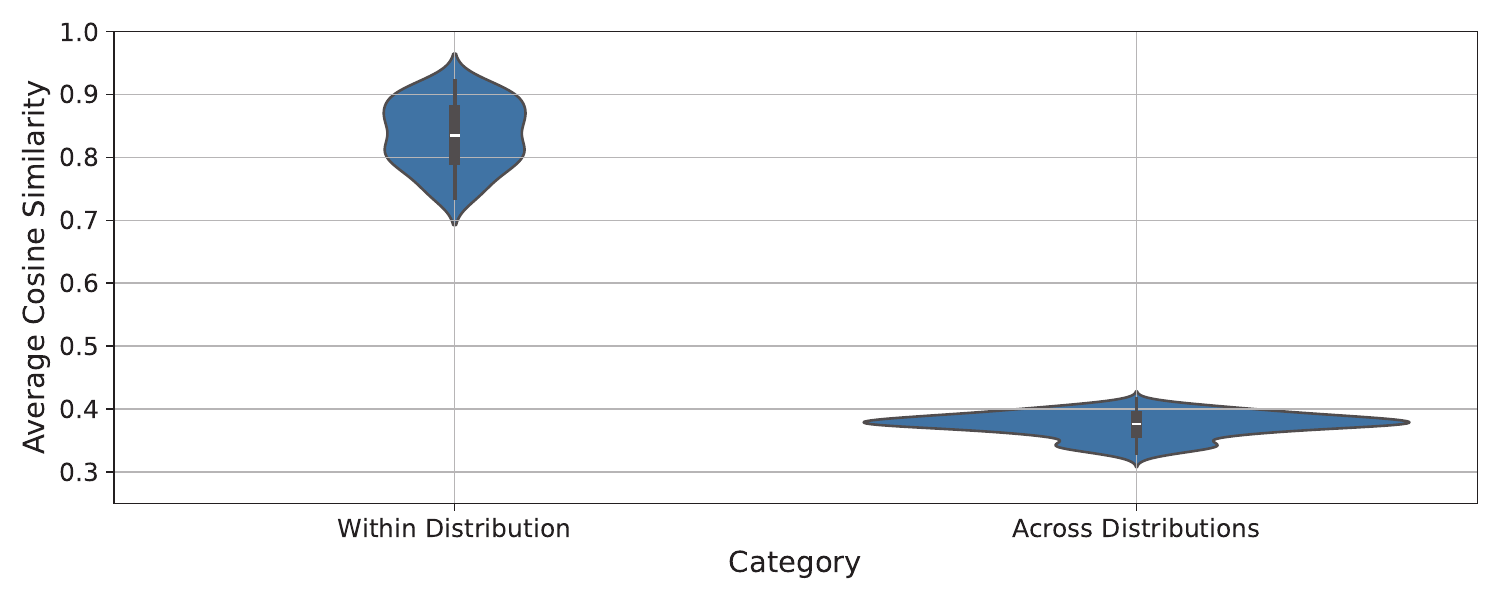}
    \includegraphics[width=1.0\columnwidth]{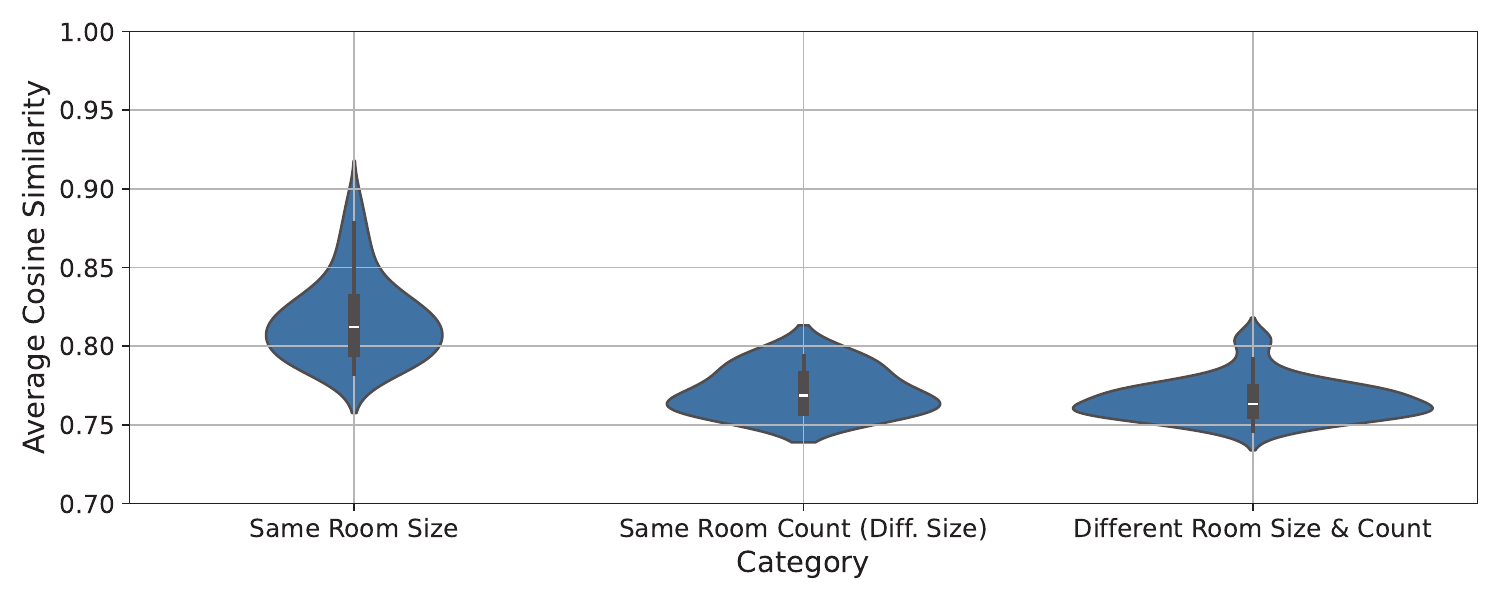}
    \includegraphics[width=1.0\columnwidth]{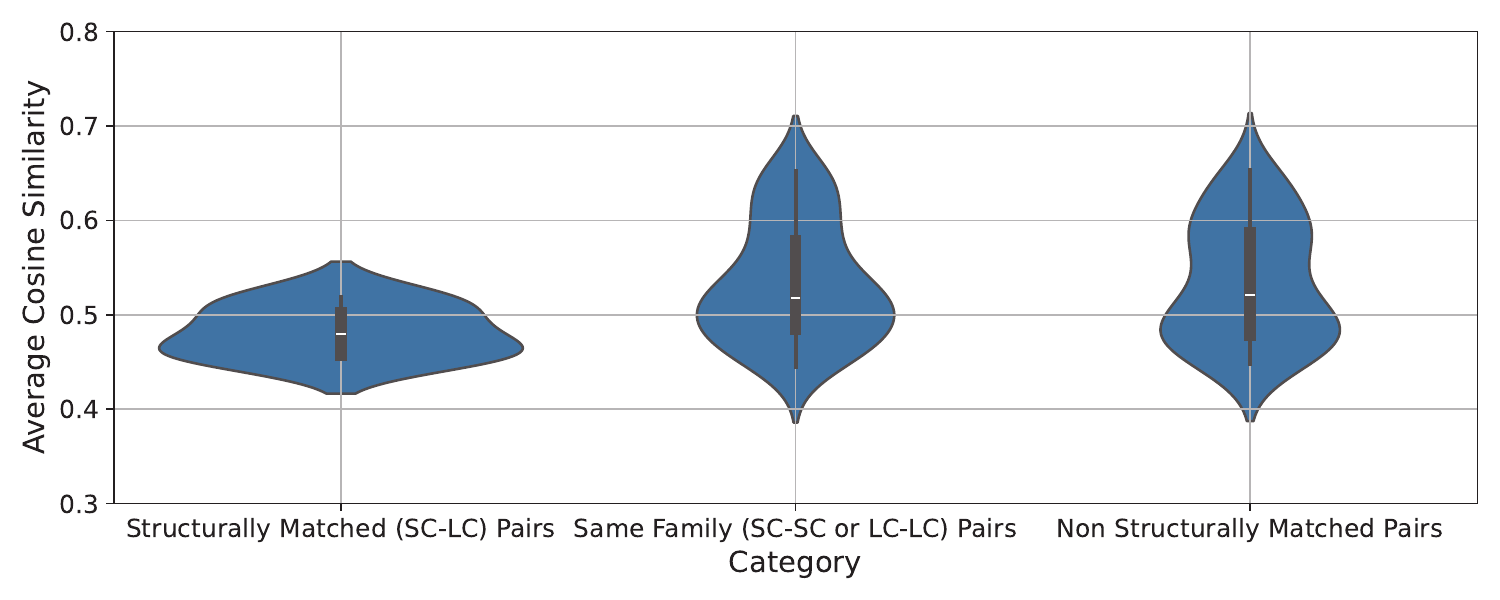}
    \caption{Violin plot showing the distribution of pairwise cosine similarities between task embeddings in the image sequence learning benchmark (top), MiniHack MultiRoom (middle), and MiniGrid Crossing (bottom).}
    \label{fig:violin_mctgraph}
\end{figure}

Figure~\ref{fig:violin_mctgraph}(top) shows the distributions of pairwise cosine similarities between groups of tasks. Pairs are grouped as either \textit{Within Distribution}, where both tasks share the same image dataset and trajectory (e.g., Dataset 1-Dataset 1), or \textit{Across Distributions}, where tasks come from different datasets and trajectories (e.g., Dataset 1–Dataset 3). Tasks within the same distribution show significantly higher similarity, indicating that MOSAIC embeddings reflect shared perceptual features and curriculum alignment.
The MiniHack benchmark extends the MiniGrid MultiRoom curriculum. It consists of tasks ranging from two to eight rooms in size, with task IDs 0–6 corresponding to room size \(4 \times 4\) and tasks 7–13 to size \(6 \times 6\). Canonical pairs are defined by room count; for example, task 0 and task 7 are both 2-room versions of different sizes. Cosine similarity analysis reveals clear grouping by room size, with tasks within the same spatial resolution exhibiting stronger similarity than canonical pairs across sizes, indicating that visual and navigational complexity induced by room dimensions plays a more significant role in shaping the embedding space than shared structural depth alone.

The matrix \(\hat{\beta}\) shows that knowledge reuse in MiniHack occurs predominantly within the same room size group. Smaller tasks, such as the 2-room and 3-room configurations, are reused more frequently and earlier in training, consistent with their faster convergence. As in MiniGrid, structural pairing alone does not guarantee high reuse. Figure~\ref{fig:violin_mctgraph}(middle) shows that canonical pairs (e.g., 0–7, 1–8) have only moderate similarity, while many within-group pairs yield stronger transfer potential. The dominance of reuse within room size groups may also be explained in part by the timing of performance gains. Tasks with smaller rooms converge earlier and thus become eligible for reuse earlier, which reinforces their influence even when structurally similar tasks emerge later in training.
\begin{figure*}
    \centering
    \includegraphics[width=1.0\textwidth]{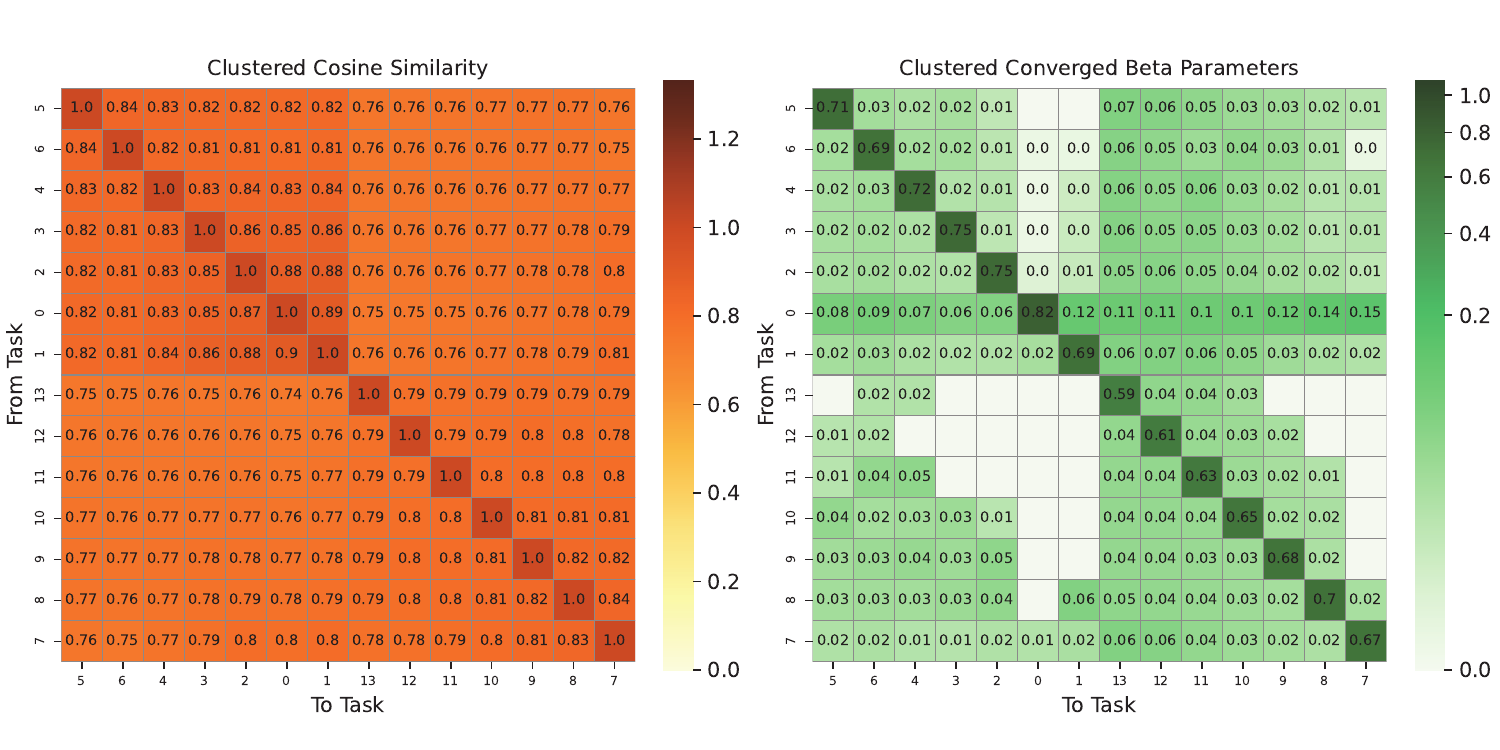}
    \caption{Pairwise (Left) cosine similarity and (Right) converged beta parameters results from the MiniHack experiment. Results are clustered using the WPGMA method.}
    \label{fig:minihack_heatmap}
\end{figure*}
In the MiniGrid benchmark, tasks are divided into two families: SimpleCrossing (SC; tasks 0–6) and LavaCrossing (LC; tasks 7–13). Each SC task has a corresponding LC task with a matched structural layout but different obstacles (walls versus lava), forming seven pairs of canonical tasks (e.g. 0-7, 1-8) that are structurally similar. Surprisingly, the pairwise cosine similarity heatmap reveals that these canonical pairs do not generally have the highest similarity. Instead, many within-family pairs show stronger similarity, particularly among SC tasks. Many non-structurally matched pairs also exhibit unexpectedly high similarity. These results suggest that actual knowlege reuse in MOSAIC is non-trivial. Rather MOSAIC agents likely reuse policies with relevant sub-trajectories that encoded in the Wasserstein task embeddings.

Figure~\ref{fig:violin_mctgraph}(bottom) categorizes task pairs into three groups: \textit{Structurally Matched Pairs (SC–LC)}, consisting of canonical SimpleCrossing–LavaCrossing pairs with identical map layouts; \textit{Same Task Type (SC–SC or LC–LC)}, where both tasks share common states but differ in layout; and \textit{Non-Structurally Matched Pairs}, which includes all remaining SC–LC task pairs not matched by structure. Structurally matched pairs do not show higher similarity than within-type pairs, suggesting that embedding similarity is more influenced by task dynamics than map layout.

\begin{figure*}
    \centering
    \includegraphics[width=1.0\textwidth]{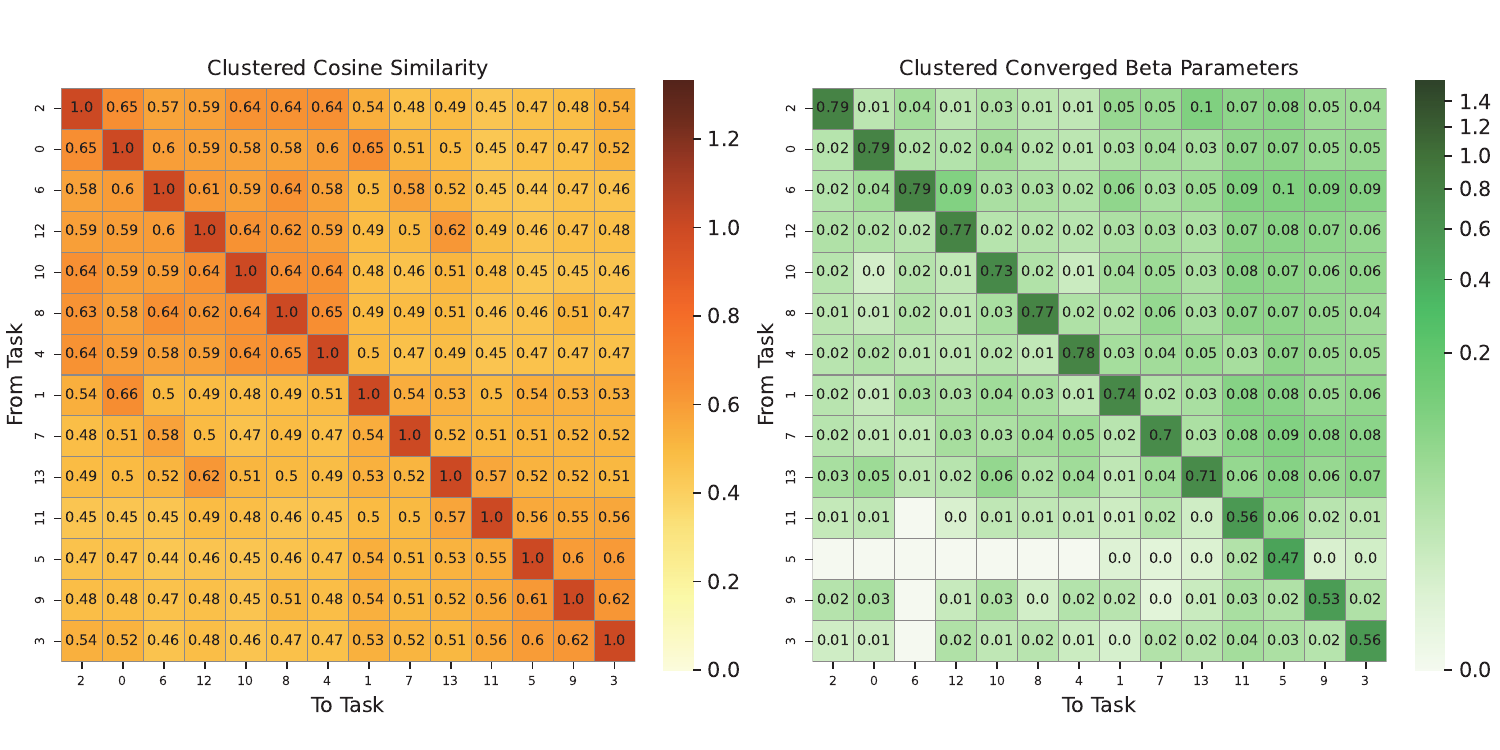}
    \caption{Pairwise (Left) cosine similarity and (Right) converged beta parameters results from the MiniGrid experiment. Results are clustered using the WPGMA method, revealing two distinct distributions. Task IDs 0-6 correspond to the SimpleCrossing. Tasks 7-13 correspond to the LavaCrossing. The results indicate that the cosine similarity of Wasserstein task embeddings is able to uncover meaningful relationships.}
    \label{fig:minigrid_heatmap}
\end{figure*}

The corresponding \(\hat{\beta}\) matrix (Figure~\ref{fig:minigrid_heatmap}) indicates that knowledge reuse in MiniGrid is selective and is not determined solely by the canonical structure. Reuse tends to favor structurally simple or early-learned tasks rather than those with the highest embedding similarity. Figure~\ref{fig:violin_mctgraph}(bottom) confirms this pattern: canonical pairs show lower average similarity than both within-group and non-structurally-matched pairs, suggesting that MOSAIC's embedding space captures aspects of task dynamics that go beyond explicit structural symmetry. Reuse decisions are shaped not only by similarity but also by policy quality and learning progress. This pattern may also reflect the temporal bias in reuse. Tasks such as SC0, which reach high performance early in training, become preferred sources of knowledge. Structurally matched tasks that appear later in training are often underutilized, due to the performance-based selection mechanism (criterion 2) that filters knowledge transfer based on relative performance.
\section{Implementation Details}
\label{sec:implementation_details}
In this section we provide implementation details for the MOSAIC algorithm. MOSAIC computes embeddings from batches of state-action-reward (SAR) data collected under the current policy. These embeddings evolve over time as the policy improves. Knowledge reuse is governed by a linear combination of task-specific masks selected using similarity and performance heuristics. The linear combination is weighted by learned \(\beta\) parameters which are re-initialized at each communication event, and then fine-tuned again to reflect downstream utility to the current task.

Algorithm~\ref{alg:mosaic-forwardpass} provides pseudocode for how MOSAIC computes the forward pass using a linear combination of masks and the binarization step. Algorithm~\ref{alg:mosaic-trainer} provides an outline of the main algorithmic pipeline. The agent additionally, runs two parallel threads to run and maintain interaction communication module and model while maintaining asynchronous operation. This enables MOSAIC to learn and communicate concurrently. Pseudocode for these threads are provided in Algorithms~\ref{alg:mosaic-update} and \ref{alg:mosaic-comm}. The server to accept incoming communications is started as a parallel process inside the communication module, described in Algorithm~\ref{alg:mosaic-server}. Queries and masks are shared around the system via multi-processing queues \(Q_{\text{emb}}\) and \(Q_{\text{mask}}\). Algorithm~\ref{alg:mosaic-wdm} shows the computation of the Wasserstein task embeddings. 
\begin{table}[ht]
    \centering
    \begin{tabular}{ll}
    \toprule
    \multicolumn{2}{l}{\textbf{System 1}} \\ 
    OS & Ubuntu 22.04.5 LTS \\
    CPU & AMD EPYC 9634 (168 logical cores) \\
    GPU(s) & 2× NVIDIA A100 80GB PCIe \\
    GPU Driver & 535.216.03 \\
    CUDA Version & 12.2 \\
    RAM & 512 GB \\
    \midrule
    \multicolumn{2}{l}{\textbf{System 2}} \\ 
    OS & Ubuntu 22.04.5 LTS \\
    CPU & AMD EPYC 7713P (128 logical cores) \\
    GPU(s) & 2× NVIDIA A100 40GB PCIe \\
    GPU Driver & 535.216.03 \\
    CUDA Version & 12.2 \\
    RAM & 256 GB \\
    \midrule
    \multicolumn{2}{l}{\textbf{System 3}} \\ 
    OS & Ubuntu 22.04.5 LTS \\
    CPU & AMD Ryzen 9 5950X (32 logical cores) \\
    GPU(s) & 1× NVIDIA GeForce RTX 4090 \\
    GPU Driver & 535.183.01 \\
    CUDA Version & 12.2 \\
    RAM & 128 GB \\
    \bottomrule
    \end{tabular}
    \caption{Compute resources used to run the experiments presented in this paper for MOSAIC and baselines. We made use of the MiG configurations that are supported by the NVIDIA A100 GPUs to run each MOSAIC agent with its own instance. This configuration limits the effects of one agent on another from a hardware perspective.}
    \label{tbl:compute_resources}
\end{table}
\begin{table*}
\centering
\begin{tabularx}{\textwidth}{l|XX}
\toprule
\textbf{Module} & \textbf{Hyperparameter} & \textbf{Value (ISL, MiniGrid, MiniHack)} \\
\midrule
PPO
  & Learning rate       & \multicolumn{1}{l}{0.00025} \\
  & Backbone seed       & \multicolumn{1}{l}{9157} \\
  & Number of workers        & \multicolumn{1}{l}{1} \\
  & Discount factor     & \multicolumn{1}{l}{0.99} \\
  & GAE Tau             & \multicolumn{1}{l}{0.99} \\
  & Entropy             & \multicolumn{1}{l}{0.01} \\
  & Rollout length      & 512 (ISL), 2048 (MiniGrid, MiniHack) \\
  & Optimization iterations     & \multicolumn{1}{l}{8} \\
  & Mini-batches        & \multicolumn{1}{l}{64} \\
  & PPO ratio clip      & \multicolumn{1}{l}{0.1} \\
  & Gradient clip       & \multicolumn{1}{l}{5} \\
  & Total training steps & 102,400 (ISL), 1,024,000 (MiniGrid), 2,048,000 (MiniHack) \\
\midrule
Comm.
  & Query frequency (iterations) & 10 (ISL, MiniHack), 25 (MiniGrid) \\
  & Query wait (ms)     & \multicolumn{1}{l}{0.3} \\
  & Mask wait (ms)      & \multicolumn{1}{l}{0.3} \\
\midrule
Embedding
  & Reference size (\(M\))   & \multicolumn{1}{l}{50} \\
  & Number of samples (\(N\))     & \multicolumn{1}{l}{128} \\
  & Frequency (iterations)   & \multicolumn{1}{l}{1} \\
\bottomrule
\end{tabularx}
\caption{Hyperparameters used in the experiments presented in this paper. Values are provided for the reinforcement learning algorithm, the communication module and embedding module.}
\label{tbl:all_hyperparameters}
\end{table*}
\begin{table}
\centering
\begin{tabularx}{\columnwidth}{l l}
\toprule
\textbf{Package/Framework} & \textbf{Version} \\
\midrule
Python           & 3.9.13    \\
PyTorch          & 2.4.0     \\
TorchVision      & 0.19.0    \\
TorchAudio       & 2.4.0     \\
NumPy            & 1.24.3    \\
SciPy            & 1.10.1    \\
Pandas           & 2.2.2     \\
Scikit-learn     & 1.5.1     \\
Gym              & 0.26.0    \\
Gym-Minigrid     & 1.2.2     \\
Gym-CTGraph      & 0.1 (dev) \\
OpenCV-Python    & 4.10.0.84 \\
Matplotlib       & 3.9.2     \\
Seaborn          & 0.13.2    \\
TensorBoardX     & 2.6.2.2   \\
POT (Optimal Transport) & 0.9.4 \\
CUDA Toolkit     & 12.1.105  \\
cuDNN            & 9.1.0     \\
\bottomrule
\end{tabularx}
\caption{Software packages and frameworks used for reproducibility.}
\label{tbl:software_environment}
\end{table}
\begin{table}
\centering
\begin{tabularx}{\columnwidth}{l X}
\toprule
\textbf{Network} & \textbf{Architecture Details} \\
\midrule
\textbf{Fully Connected} & 
\textbf{Feature extractor:} Linear (input, 200) $\rightarrow$ Linear (200, 200) $\rightarrow$ Linear (200, 200) \newline
\textbf{Critic:} Linear (200, 1) \newline
\textbf{Actor:} Linear (200, action) \\
\textbf{Convolutional} & 
\textbf{Feature extractor:} Conv (input, 16, kernel 8, stride 4) $\rightarrow$ Conv (16, 32, kernel 4, stride 2) $\rightarrow$ Linear (8192, 256) \newline
\textbf{Critic:} Linear (256, 1) \newline
\textbf{Actor:} Linear (256, action) \\
\bottomrule
\end{tabularx}
\caption{Network architectures used in experiments. Image sequence learning and MiniGrid benchmarks were run using the full connected network. MiniHack experiments were run using the convolutional network}
\label{tbl:network_architectures}
\end{table}
\begin{algorithm*}
\caption{Agent Training Loop in MOSAIC}
\label{alg:mosaic-trainer}
\begin{algorithmic}[1]
\Require Assigned task \(\tau \in \mathcal{T}\), shared backbone \(\Phi\)
\Require Communication event interval \(\Delta_{\text{comm}}\), embedding interval \(\Delta_{\text{embed}}\)
\Require Replay buffer \(\mathcal{D}_\tau\), rollout length \(z\), total steps \(\Omega\), cosine similarity threshold \(\theta = 0.5\), \(N\gets128\)
\State Initialize mask \(\phi_\tau\), embedding \(v_\tau \gets \mathbf{0}\), return estimate \(r \gets 0\), step counter \(t \gets 0\)
\State Initialize peer mask set \(P_0 = \emptyset\), and linear combination parameters \({\beta} \in \mathbb{R}^{1 + |P_0|}\)
\State Launch \textbf{parallel threads}:
\State \hspace{1em} \textsc{ReplacePeerMasks}($Q_{\text{mask}}$) \hfill \Comment{Receives and applies mask updates Alg.~\ref{alg:mosaic-update}}
\Statex \hspace{1em} \textsc{CommunicationModule}($Q_{\text{emb}}, Q_{\text{mask}}$) \hfill \Comment{Handles queries and peer responses Alg.~\ref{alg:mosaic-comm}}
\While{\(t < \Omega\)}
    \State Roll out policy \(\pi_\tau = \pi_{\Phi \odot g(\phi_\tau^\text{lc})}\) for \(z\) steps\Comment{Forward pass Alg.~\ref{alg:mosaic-forwardpass}}
    \State Store transitions in \(\mathcal{D}_\tau\); update iteration return estimate \(r\)
    \State Update \(\phi_\tau\) using policy gradient on loss \(\mathcal{L}(\pi_\tau)\) \Comment{Via Eq.~\ref{eq:linearcombination}}
    \State \(t \gets t + z\)
    \If{\(t \bmod \Delta_{\text{embed}} = 0\) \textbf{and} \(|\mathcal{D}_\tau| \ge N\)}
        \State Sample \(N\) samples from \(\mathcal{D}_\tau\) and compute task embedding \(v_\tau \gets \psi(\mu_\tau)\)\Comment{Alg.~\ref{alg:mosaic-wdm}}
    \EndIf
    \If{\(t \bmod \Delta_{\text{comm}} = 0\)}
        \State Send \texttt{TEQ} for \((v_\tau, r)\) to \textsc{CommunicationModule} via \(Q_{\text{emb}}\) and acquire set of peer masks \(P_{c+1}=\{\phi_1...\phi_K\}\) in \textsc{CommunicationModule}
        \State Update model with \(P_{c+1}\) for forward pass via \textsc{ReplacePeerMasks} through \(Q_\text{mask}\)
    \EndIf
    \If{\(t=\Omega\)}
        \State Enter idle mode to keep \textsc{CommunicationModule} active
    \EndIf
\EndWhile
\end{algorithmic}
\end{algorithm*}
\begin{algorithm*}[ht]
\caption{Updating set of peer masks and resetting linear combination parameters}
\label{alg:mosaic-update}
\begin{algorithmic}[1]
\Require latest iteration return \(r\)
\Procedure{ReplacePeerMasks}{$Q_{\text{mask}}$}
    \While{running}
        \State Wait for set of new peer masks \(P_{c+1}=\{\phi^\text{new}_1,\dots,\phi^\text{new}_K\}\) from \(Q_\text{mask}\)
        \State Update \(\phi_\tau \gets \beta_\tau \phi_\tau + \sum_{k \in P_{c}} \beta_k \phi_k\)\Comment{Consolidation step}
        \State Cache consolidated mask for forward pass
        \State Update \(\beta_\tau \gets 0.5 + 0.5r\), \(\beta_k\gets\frac{0.5(1-r)}{|P_{c+1}|} \quad \forall k \in P_{c+1}\)
        \State \(P_c \gets P_{c+1}\)
    \EndWhile
\EndProcedure
\end{algorithmic}
\end{algorithm*}
\begin{algorithm*}[ht]
\caption{{MOSAIC Communication Module}}
\label{alg:mosaic-comm}
\begin{algorithmic}[1]
\Require Agent's IP/Port \(\texttt{addr}_i\), Communication graph \(\mathcal{G}=\{\mathcal{I},\mathcal{E}\}\), where \(\mathcal{E} = \{(i, j) \mid i, j \in \mathcal{I},\ i \ne j\}\)
\Procedure{CommunicationModule}{$Q_{\text{emb}}, Q_{\text{mask}}$}
    \State Initialize \textbf{parallel process} \textsc{CommunicationServer(\(Q_{\text{mask}}\))}\Comment{Begins server process Alg.~\ref{alg:mosaic-server}}
    \While{running}
        \State Wait for query \((v_i, r_i, \texttt{addr}_i)\) from \(Q_{\text{emb}}\)
        \State Send query to known peers
        \State Wait for set of peer \texttt{QR}s \(\{(v_j, r_j, \text{mask\_id}_j, \texttt{addr}_j)\}\) from \textsc{CommunicationServer}
        \State Select peers \(P \gets \{j\ |\ \mathbb{I}_{\text{align}} \text{ and } \mathbb{I}_\text{perf}\}\)
        \State Send mask request (\texttt{MR}) to selected peers \(P\)
        \State Wait for peer masks from \textsc{CommunicationServer} and send to model via \(Q_{\text{mask}}\)
    \EndWhile
\EndProcedure
\end{algorithmic}
\end{algorithm*}
\begin{algorithm*}[ht]
\caption{Forward pass using masks for network layer \(l\)}
\label{alg:mosaic-forwardpass}
\begin{algorithmic}[1]
\Require Masks \(\phi_\tau\), peer mask set \(P=\{\phi_k\}_{k=1}^{|P|}\), parameters \(\beta_\tau\), \(\{\beta_k\}_{k=1}^{|P|}\), backbone \(\Phi\)
\Procedure{ForwardPass}{\textbf{x}}
    \If{\(|P| = 0\)} \Comment{no peer masks available}
        \State Set \(\phi_\tau^\text{lc}\gets\phi_\tau\)
    \Else
        \State \(\phi_\tau^\text{lc}\gets g\left(\beta_\tau\phi_\tau+\sum_{k\in P}\beta_k\phi_k\right)\) \Comment{Eq.~\ref{eq:linearcombination}}
    \EndIf
    \State \(\pi_\tau \gets \pi_{\Phi\odot\phi_\tau^\text{lc}}\)\Comment{Modulate backbone with LC mask to get policy}
    \State \(y \gets \pi_\tau\cdot \textbf{x}\)\Comment{Compute output from policy}
    \State \Return \(y\)
\EndProcedure
\end{algorithmic}
\end{algorithm*}
\begin{algorithm*}[ht]
\caption{{Communication server}}
\label{alg:mosaic-server}
\begin{algorithmic}[1]
\Require World size \(W\), Agent's IP/Port \(\texttt{addr}_i\)
\Procedure{CommunicationServer}{$Q_{\text{mask}}$}
    \State Bind TCP socket to \((\texttt{addr}_i)\)
    \State Listen for connections with backlog size \(W\)
    \While{running}
        \State Accept incoming connection \((\texttt{conn}, \texttt{remote\_addr})\)
        \State Receive message from connection
        \If{message is an \texttt{TEQ}}
            \If{Server-side selection}
                \State Select mask(s) and respond with \texttt{MTR} to \texttt{remote\_addr}
            \Else
                \State Respond to \texttt{remote\_addr} with embedding and performance (\texttt{QR})
            \EndIf
        \ElsIf{message is an \texttt{QR}}
            \State Store \texttt{QR} \((v_j,r_j,\text{mask\_id},\texttt{addr}_j)\) in set of query responses
        \ElsIf{message is an \texttt{MR}}
            \State Fetch corresponding mask for mask id and respond to \texttt{remote\_addr} with \texttt{MTR}
        \ElsIf{message is an \texttt{MTR}}
            \State Add mask to set of acquired peer masks
        \EndIf
    \EndWhile
\EndProcedure
\end{algorithmic}
\end{algorithm*}
\begin{algorithm*}[h]
\caption{Wasserstein Embedding Computation}
\label{alg:mosaic-wdm}
\begin{algorithmic}[1]
\Procedure{ComputeEmbedding}{$\mathcal{D}$}
    \State Compute embedding \(v_\tau^{\text{new}} \gets \psi(\mathcal{D})\) \Comment{via Eq.~\ref{eq:embedding_mapping}}
    \State \(v_\tau \gets \frac{1}{2}(v_\tau + v_\tau^{\text{new}})\) \Comment{Moving average update}
    \State \textbf{return \(v_\tau\)}
\EndProcedure
\end{algorithmic}
\end{algorithm*}
\end{document}